# Information Coordination as a Bridge: A Neuro-Symbolic Architecture for Reliable Autonomous Driving Scene Understanding

Shuo Liu[a], Lei Shi[b,c], Haowen Liu[a], Jing Xu[a], Yufei Gao[b,c], Yucheng Shi*[a]

[a] *School of Computer and Artificial Intelligence, Zhengzhou University, 450001, Zhengzhou, China*

[b] *School of Cyber Science and Engineering, Zhengzhou University, 450001, Zhengzhou, China*

[c] *Songshan Laboratory, 450003, Zhengzhou, China*

**Abstract**

Reliable autonomous driving requires scene understanding that is semantically consistent across heterogeneous sensors and verifiable at the reasoning stage. However, many recent LLM-driven driving systems attach the language model as a post-processor and force it to reason over redundant or conflicting perception outputs, which can amplify hallucinated entities and unsafe conclusions. This paper proposes InfoCoordiBridge, a BEV-centric neuro-symbolic architecture that inserts an explicit coordination bridge between perception and language reasoning. InfoCoordiBridge comprises (i) a unified multi-agent perception layer that outputs typed structured facts together with modality-focused synopses, (ii) an ICA module that aligns and fuses multi-source outputs into a single SceneSummary, and (iii) an SSRE module that performs SceneSummary-grounded reasoning with verification. Experiments on nuScenes and Waymo show that ICA preserves competitive 3D detection accuracy while substantially improving fusion consistency, reducing redundancy to below 1% and achieving about 98% attribute agreement. On NuScenes-QA and a template-aligned Waymo-QA benchmark, SSRE improves factual grounding and reduces hallucinated entity mentions compared with representative VLM and agentic baselines. Overall, by coordinating multi-sensor outputs into a single conflict-aware SceneSummary before prompting, InfoCoordiBridge prevents redundant and cross-modally inconsistent perception evidence from propagating into high-level reasoning.

## 1. Introduction

Autonomous driving robustness critically depends on whether the perception stack can deliver accurate and interpretable scene understanding in complex, open-world traffic. Multimodal sensor fusion has therefore become a cornerstone for reliable environment perception (Peng, et al., 2022). Among existing representations, bird's-eye-view (BEV) unifies heterogeneous sensor viewpoints while retaining both rich visual semantics and precise geometric structure. Representative BEV-based fusion methods, such as BEVFusion (Z. Liu, et al., 2023) and TransFusion (Bai, et al., 2022), show that feature-level fusion in BEV space can substantially improve downstream tasks (e.g., 3D object detection), marking a key milestone in multimodal perception.

Despite these advances, BEV-centric fusion remains largely confined to a closed-set perception paradigm: models typically output a predefined set of low-level annotations such as bounding boxes and segmentation masks (L. Wang, et al., 2023). However, high-level driving decisions require deeper scene understanding—inferring traffic participants' intents, resolving complex interactions, and reasoning about atypical events—which calls for open-world commonsense and logical reasoning. Large language models (LLMs) have demonstrated strong capabilities in natural language understanding and reasoning, offering a promising opportunity to bridge the semantic gap between perception outputs and decision making (Ahn, et al., 2022).

Motivated by this opportunity, recent studies integrate LLMs into autonomous driving pipelines. They condition planning on perception outputs (X. Huang, et al., 2024) or answer natural-language questions about driving scenes (Qian, et al., 2024). This trend, however, exposes a critical tension: cross-modal perception results are often semantically inconsistent, while LLMs lack built-in priors to reconcile such inconsistencies and are therefore prone to hallucinations. Prior work reports that, in safety-critical domains such as autonomous driving, LLMs may generate plausible-sounding yet reality-inconsistent conclusions (Chakraborty, et al., 2025; Heo, et al., 2025). The resulting risk has become a major bottleneck for trustworthy open-world scene understanding and directly motivates this work.

At the architectural level, many pioneering attempts treat the LLM as a loosely coupled post-processor placed after the perception front-end. Under this paradigm, inconsistencies in upstream perception cannot be corrected before decision-making and are directly propagated into high-level reasoning. The LLM therefore consumes perception outputs as fixed inputs, without an explicit mechanism to reconcile conflicts prior to reasoning. When camera and LiDAR reports conflict due to noise, the pipeline lacks an explicit step to detect the inconsistency, match the corresponding entities, and reconcile their attributes. This paradigm inevitably amplifies hallucination risks and renders the decision pipeline opaque, posing severe challenges for safety-critical applications.

Taken together, semantic inconsistency and hallucination-prone reasoning constitute key obstacles to reliable driving cognition. The central challenge is therefore to design an autonomous driving cognition framework that leverages the reasoning power of LLMs while ensuring that decisions are grounded in an accurate, globally consistent world representation.

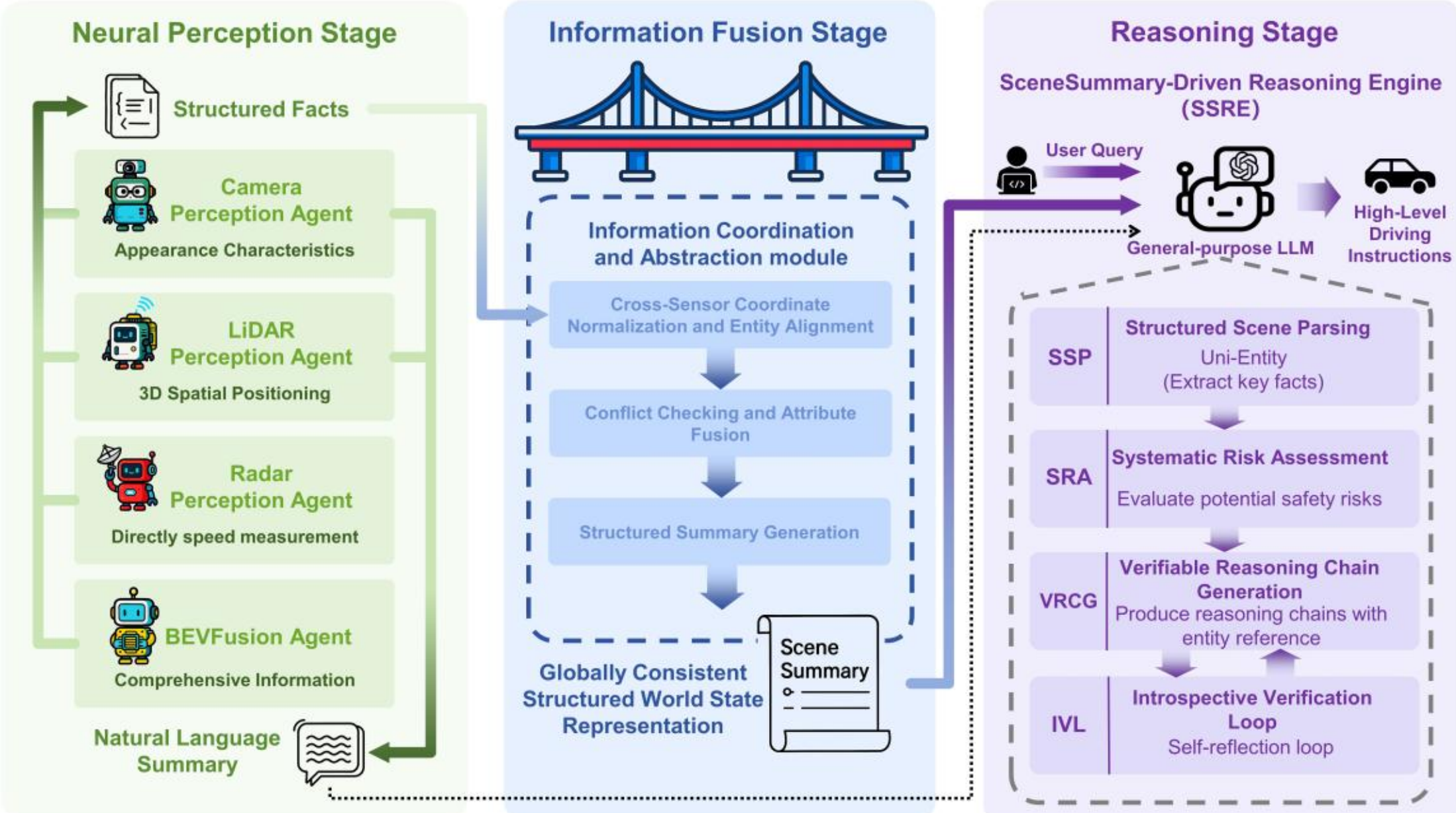

Figure 1. Overview of InfoCoordiBridge: an explicit coordination bridge from multimodal perception to verifiable reasoning.

To address this challenge, we propose InfoCoordiBridge, a structured neuro-symbolic architecture that introduces an Information Coordination and Abstraction (ICA) module as an explicit interface between perception and reasoning. As illustrated in Figure 1, the architecture comprises:

(i) a unified-interface multi-agent perception layer that outputs standardized structured facts (JSON) together with brief modality-focused synopses, where specialized agents (Camera, LiDAR, Radar, and BEVFusion) contribute complementary evidence under the same schema. This standardization reduces redundant descriptions and stabilizes the inputs to downstream coordination and reasoning.

(ii) an ICA module that performs deterministic coordinate normalization, cross-source entity alignment, and conflict-aware attribute fusion to generate a single auditable SceneSummary with traceable source references. This module helps fix the direct propagation of duplicated entities and cross-modal inconsistencies into the reasoning stage.

(iii) a SceneSummary-Driven Reasoning Engine (SSRE) that executes a four-stage verifiable workflow (SSP, SRA, VRCG, and IVL) to keep LLM inference entity-grounded and iteratively checked. This workflow helps improve hallucination-prone reasoning and reduces unverified safety conclusions under open-world uncertainty.

Building on this architecture, our contributions are threefold:

(1) We propose InfoCoordiBridge, a BEV-centric neuro-symbolic framework that explicitly bridges perception and language reasoning via information coordination.

(2) We introduce the ICA-based SceneSummary construction for deterministic multi-source alignment and conflict-aware fusion with auditable traces.

(3) We develop SSRE as a verifier-aware reasoning scheme over SceneSummary and empirically demonstrate improved consistency and reduced hallucinations on multi-sensor benchmarks and driving QA tasks.

The remainder of the paper is organized as follows. Section 2 reviews related work. Section 3 formalizes the task definition. Section 4 details the proposed method, including the multi-agent perception layer, the ICA module, and the SSRE workflow. Section 5 presents experimental results and discussion. Section 6 concludes the paper and outlines future directions.

## 2. Related work

Autonomous driving research is gradually extending from perception fusion and decision learning to cognition-level reasoning and semantic validation in open environments. To clarify the connections between prior studies and our framework, we categorize related work into three threads: (i) BEV-based multimodal perception and fusion, (ii) applications of vision-language models (VLMs) and large language models (LLMs) in autonomous driving, and (iii) verifiable neuro-symbolic reasoning with formal safety constraints. These threads correspond to the key components of our approach—an information coordination bridge, a layered neuro-symbolic architecture, and a SceneSummary-Driven Reasoning Engine (SSRE).

### 2.1 BEV-based multimodal perception and fusion

Multimodal perception is a cornerstone of autonomous driving, aiming to fuse heterogeneous sensor observations into a unified spatial representation that supports consistent environment understanding. BEVFusion (Z. Liu, et al., 2023) pioneered feature-level alignment in the bird's-eye-view (BEV) space, effectively mitigating geometric discrepancies across sensors. Subsequent methods such as BEVDepth (Y. Li, et al., 2023) and BEVFormer (Z. Li, et al., 2024) introduced depth-aware modeling and temporal attention to enhance spatiotemporal consistency and robustness. More recent approaches, including PolarFormer (Jiang, et al., 2023), FB-BEV (Z. Li, et al., 2023), and TransFusion (Bai, et al., 2022), further leverage Transformer-based designs to optimize cross-modal semantics, consolidating BEV as a prevailing representation for multi-sensor fusion. Complementary to these BEV-centric pipelines, strong modality-specific baselines (e.g., CenterPoint (Yin, et al., 2021) for LiDAR and RoPETR (Ji, et al., 2025) for multi-view cameras) are often reported to contextualize the performance limits of each sensor stream. Recent end-to-end camera–LiDAR fusion detectors, such as Fully Sparse Fusion (FSF) (Y. Li, et al., 2024), GraphBEV (Song, et al., 2024) and HS-Fusion (Tang, et al., 2024), further investigate sparse cross-modal interaction and robust BEV feature alignment under challenging sensor discrepancies.

Despite strong performance on standard perception tasks (e.g., 3D object detection and tracking), BEV-based fusion methods often provide limited semantic abstraction and a task-specific interface to downstream reasoning. Their outputs typically remain at geometric or dense-prediction levels (e.g., 3D boxes or segmentation), lacking an interpretable and structured semantic interface that can directly support high-level reasoning and planning. Motivated by this limitation, our framework introduces an explicit information coordination module that unifies coordinates, entities, and attributes across modalities and compiles a structured SceneSummary, which serves as a normalized and auditable input to downstream neuro-symbolic reasoning.

### 2.2 VLM/LLM applications in autonomous driving

With the rapid development of VLMs and LLMs, recent studies have explored integrating language models into autonomous driving to enhance semantic understanding and provide more interpretable decisions (Yang, et al., 2023). Early attempts such as DriveGPT (X. Huang, et al., 2024) and GPT-Driver (Mao, et al.) explicitly model driving intent and scene semantics through language, enabling mappings from perceptual inputs to language-conditioned decisions. DriveLM (Sima, et al., 2024), DriveVLM (Tian, et al., 2025) and Talk2Drive (Cui, et al., 2024) further demonstrate the potential of LLMs for natural-language interaction and instruction following, allowing systems to articulate driving rationales in human-readable form. Recent work also combines knowledge graphs with LLMs via RAG to deliver explainable behavior predictions (Hussien, et al., 2025). At the representation level, Talk2BEV (Choudhary, et al., 2024) augments BEV maps with language supervision to support interactive querying and interpretation, indicating the value of BEV-grounded language interfaces for driving scene understanding.

In addition, general-purpose VLMs such as LLaVA (H. Liu, et al., 2023), BLIP-2 (J. Li, et al., 2023), and Flamingo (Alayrac, et al., 2022) have been applied to visual question answering and high-level scene description tasks, offering richer semantic supervision signals. Beyond single-model usage, there is a growing trend to wrap foundation models into agentic systems. For example, DriveAgent (Hou, et al., 2025) combines LLMs with a multi-agent framework to enable collaborative reasoning via language, and PlanAgent (Zheng, et al., 2024) targets closed-loop motion planning by integrating multimodal LLMs with hierarchical reasoning and reflection components.

However, many pipelines that incorporate language models emphasize language-conditioned perception or descriptive reasoning, whereas systematic modeling of semantic consistency and verifiability remains limited. In safety-critical autonomous driving, cross-modal inconsistencies can easily propagate into hallucination-prone high-level inferences. Our layered neuro-symbolic architecture addresses this gap by placing an explicit coordination and abstraction interface between perception and reasoning. It produces a single-source SceneSummary for grounding, and SSRE further enforces entity-anchored reasoning and consistency checking to support reliable open-domain scene understanding.

### 2.3 Verifiable neuro-symbolic reasoning and formal safety constraints

Achieving explainability and safety in complex driving scenes requires reasoning processes that are verifiable. A traditional line of work relies on formal verification and model checking to certify properties of neural components (e.g., perception networks) and to evaluate system behaviors under prescribed assumptions (Elboher, et al., 2024). While such methods offer rigorous guarantees, they often depend on static modeling assumptions and can be difficult to adapt to dynamic, semantics-rich environments.

To improve logical consistency and interpretability, researchers have introduced neuro-symbolic reasoning frameworks that represent semantic relations using symbolic or probabilistic logic. For example, Maene and De Raedt (Maene & De Raedt, 2023) study soft-unification in deep probabilistic logic, and Choi et al. (Choi, et al., 2024) investigate neuro-symbolic video understanding with explicit relational structure. Nevertheless, many of these models are developed for relatively generic reasoning settings and are not directly tailored to end-to-end autonomous driving cognition.

With the emergence of LLM reasoning capabilities, explainable driving has also attracted increasing attention. DriveGPT4 constructs traceable reasoning paths through language generation (Xu, et al., 2024), and DriveLM makes dependencies among perception, prediction, and planning explicit through graph-structured question answering (Sima, et al., 2024). Although these efforts enhance transparency, they do not yet provide a unified mechanism that simultaneously reconciles perceptual uncertainty and verifies reasoning outputs against a structured source of truth.

More recently, research has shifted toward reflective and verifiable reasoning chains. ReAct (Yao, et al., 2022) and Reflexion (Shinn, et al., 2023) incorporate feedback and self-checking mechanisms to support iterative refinement. Parameshwaran and Wang (Parameshwaran & Wang, 2024) combine Linear Temporal Logic (LTL) and Signal Temporal Logic (STL) to generate and verify safety-compliant trajectories for autonomous vehicles, and TrustLLM (Y. Huang, et al., 2024) highlights benchmarks and requirements for trustworthy reasoning in high-stakes settings. Overall, a closed-loop framework that integrates (i) uncertainty reduction and conflict resolution in multimodal perception, (ii) logic-consistent entity-anchored reasoning, and (iii) verification and revision of reasoning outputs remains largely absent in autonomous driving.

Motivated by this gap, we propose a SceneSummary-Driven Reasoning Engine (SSRE) that couples neural semantic abstraction with symbolic consistency verification. By combining a verifiable reasoning chain with an introspective verification loop, SSRE enables closed-loop control from semantic understanding to reasoning validation, improving reliability and safety interpretability for autonomous driving scene understanding.

Table 1 summarizes key capability dimensions relevant to reliable scene understanding, including whether a method aggregates multiple sources, explicitly resolves conflicts, anchors reasoning to entities, performs self-reflection, and adopts an external verifier. Representative BEV fusion methods (e.g., BEVFusion and TransFusion) leverage multi-source sensing but do not explicitly reconcile semantic conflicts or verify reasoning outputs. Agent-based approaches (e.g., DriveAgent) partially address multi-source coordination, yet they often lack verification mechanisms. In contrast, our approach is designed to make conflict resolution and verification explicit at the interface and reasoning levels, enabling more auditable and reliable open-domain scene understanding.

Table 1. Comparison of the proposed method with representative approaches (✓ = Yes, ✕ = No)

| Method | Multi-Fact Source | Explicit Conflict Resolution | Entity Anchoring | Self-Reflection | External Verifier |
|---|---|---|---|---|---|
| BEVFusion | ✓ | ✕ | ✕ | ✕ | ✕ |
| TransFusion | ✓ | ✕ | ✕ | ✕ | ✕ |
| DriveAgent | ✓ | ✓ | ✓ | ✕ | ✕ |
| DriveVLM | ✕ | ✕ | ✓ | ✕ | ✕ |
| Ours (InfoCoordiBridge) | ✓ | ✓ | ✓ | ✓ | ✓ |

## 3. Task definition

We study reliable open-world scene understanding for autonomous driving. Given synchronized multimodal sensor observations, the goal is to produce (i) a globally consistent, provenance-aware world state that can serve as a single source of truth, and (ii) a high-level decision with a traceable justification grounded in that world state. To enable an auditable perception-to-reasoning pipeline, we decompose the task into three hierarchical sub-tasks: multi-agent perception, information coordination and abstraction, and SceneSummary-driven reasoning. Each sub-task outputs a well-typed artifact that is consumed by the subsequent stage.

### 3.1 Multi-agent perception

Let $X = \{x^{cam}, x^{lidar}, x^{radar}, \ldots\}$ denote synchronized raw sensor data at a timestamp. The multi-agent perception layer comprises $K$ specialized perception agents $\{\pi_k\}_{k=1}^{K}$. Each agent may consume a single modality or a multimodal subset $X_k \subseteq X$, and produces a pair of outputs: a structured fact set and an optional natural-language synopsis,

$$(F_k, A_k) = \pi_k(x_k), k = 1, \ldots K. \tag{1}$$

Here $F_k$ is a machine-readable structured fact set serialized as JSON (e.g., entity detections with attributes following a unified schema), and $A_k$ is an agent-specific natural-language synopsis generated by an embedded lightweight language model to summarize modality-unique cues. The agent design and the unified output interface are detailed in Section 4.1.

### 3.2 Information coordination and abstraction

Given the collection of structured facts $\mathrm{F} = \{F_k\}_{k=1}^{K}$, the Information Coordination and Abstraction (ICA) module deterministically compiles a globally consistent world state $S$, termed SceneSummary (Section 4.2). ICA performs cross-sensor coordinate normalization, hierarchical entity alignment, and attribute-level fusion/conflict resolution. At an abstract level, this coordination process can be viewed as minimizing the weighted discrepancy between a candidate world state and the agent-provided facts,

$$S = \arg\min_{\tilde{S}} \sum_{k=1}^{K} \omega_k d\left(T_k(F_k), g_k(\tilde{S})\right) \tag{2}$$

Here $T_k(\cdot)$ denotes the coordinate transformation that maps agent $k$ 's facts into a common ego/BEV frame, $d(\cdot,\cdot)$ is a distance metric over entities and attributes, $\omega_k$ is a dynamic reliability weight encoding modality priors, confidence, uncertainty, and cross-source consistency, and $g_k(\tilde{S})$ projects a candidate world state into the comparable representation space of agent $k$. Importantly, ICA consumes only the structured facts $F_k$ for numeric coordination; the natural-language synopses $A_k$ are not used for alignment or fusion.

The output SceneSummary $S = \{e_j\}_{j=1}^{N}$ is a typed list of fused entities. For each entity $e_j$, continuous attributes (e.g., position and velocity) are fused via reliability-weighted estimation (Eqs. (15)–(16)); correlated-source fusion uses covariance intersection when needed (Eq. (17)); categorical attributes (e.g., class) are fused via weighted voting (Eq. (18)); and ambiguous attributes are flagged by consistency gates (Eq. (19)) (Section 4.2). In addition to state estimates, the SceneSummary records provenance (sensor sources and fusion decisions) and uncertainty indicators, enabling it to serve as a single, auditable source of truth for downstream reasoning.

### 3.3 SceneSummary-driven reasoning

Given the coordinated world state $S$ produced by ICA, an optional user query $Q$, and optional auxiliary synopses $A = \{A_k\}_{k=1}^{K}$, the SceneSummary-Driven Reasoning Engine (SSRE) generates high-level decisions and explanations,

$$(D, N) = \mathcal{R}(S, Q, A) \tag{3}$$

The structured decision $D$ is serialized as JSON (e.g., *recommended_action*, *confidence*, and *supporting_entity_ids*), and the natural-language justification $N$ must ground its factual claims by explicitly referencing entities in $S$. SSRE is implemented as a constrained chain-of-prompt workflow with an introspective verification loop (Algorithm 1 in Section 4.3), which verifies and revises draft outputs against $S$ before emitting the final decision and justification.

## 4. Method

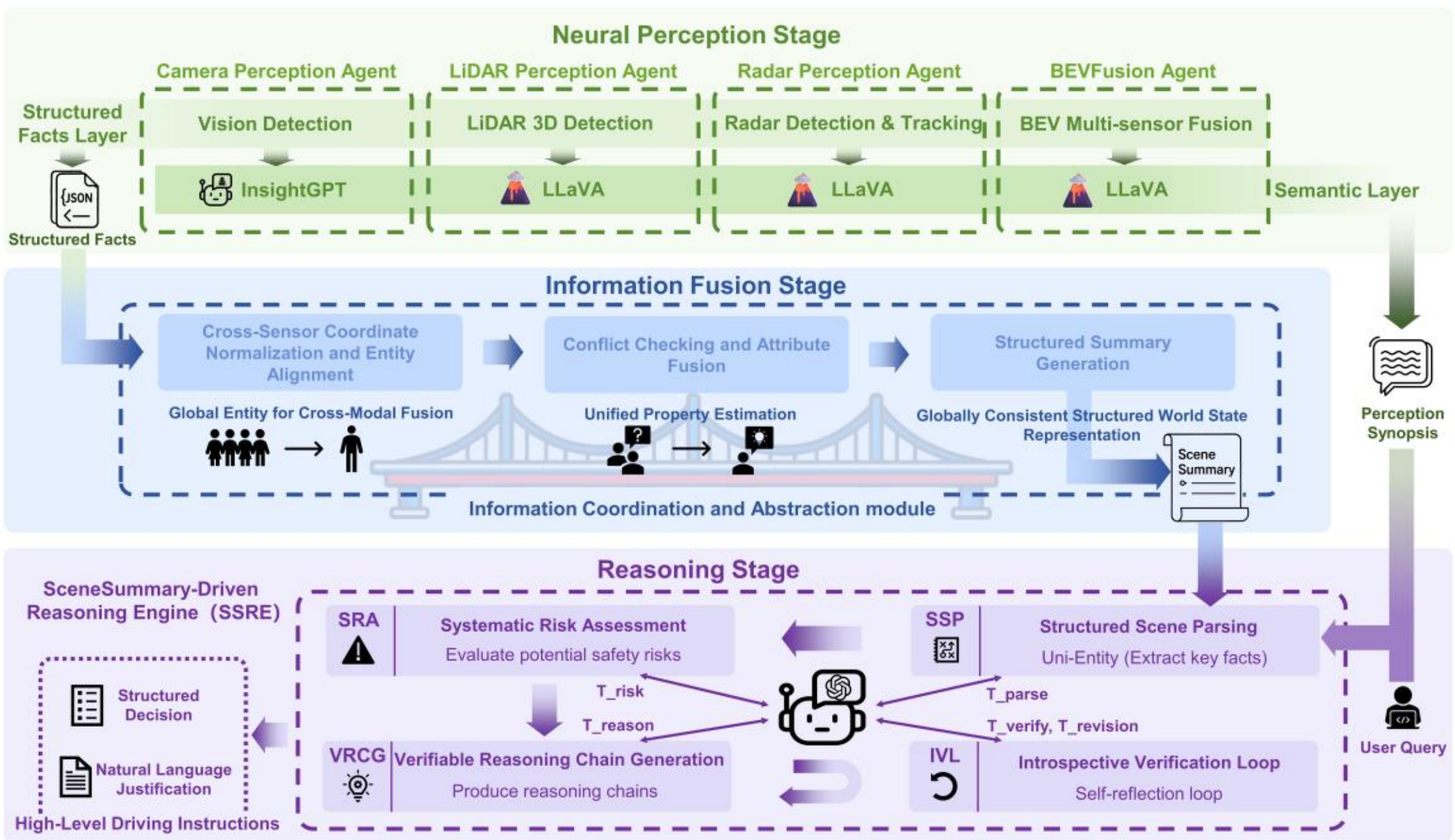

Figure 2. Architecture of InfoCoordiBridge: multi-agent perception, ICA-based coordination, and SceneSummary-driven reasoning (SSRE).

We propose InfoCoordiBridge, a BEV-centric neuro-symbolic architecture for reliable open-world autonomous driving scene understanding. The design goal is to couple (i) rich neural perception from heterogeneous sensors with (ii) deterministic symbolic coordination, so that downstream reasoning operates on a globally consistent and auditable world state rather than on redundant or conflicting perception logs.

As illustrated in Figure 2, the architecture consists of three layers. First, a multi-agent perception layer extracts complementary evidence from different sensing modalities and outputs a unified, typed interface. Second, an Information Coordination and Abstraction (ICA) layer compiles multi-source perception facts into a single, conflict-resolved world state, termed SceneSummary. Third, a SceneSummary-Driven Reasoning Engine (SSRE) performs high-level reasoning and decision making under explicit verifiability constraints, producing both a structured decision and a grounded natural-language justification.

The remainder of this section details each component. Section 4.1 describes the multi-agent perception module, including the unified output schema and synopsis protocol; Section 4.2 presents the Information Coordination and Abstraction (ICA) module; and Section 4.3 describes the SceneSummary-Driven Reasoning Engine (SSRE).

### 4.1 Multi-agent perception module

The multi-agent perception module is designed to exploit sensor complementarity while providing a disciplined interface for downstream coordination and reasoning. Instead of forcing a single monolithic model to infer all properties, we deploy multiple specialized agents—one multi-modal BEV fusion agent and several single-modal agents—to extract modality-unique evidence with minimal redundancy. Each agent outputs (i) structured, machine-readable facts $F_k$ in a unified JSON schema and (ii) a concise natural-language synopsis $A_k$ that emphasizes the agent's distinctive contribution.

Notably, the structured facts $F_k$ are the only inputs used for numeric alignment and fusion in ICA, whereas the synopses $A_k$ serve as optional auxiliary context for SSRE (e.g., for compact semantic cues that are difficult to encode as fields). This separation ensures that cross-sensor coordination remains deterministic and auditable, while high-level reasoning can still benefit from modality-specific semantics in a controlled manner.

Table 2 summarizes the four agents used in our implementation: a BEVFusion agent for global BEV-space fusion, a camera agent for appearance-centric semantics, a LiDAR agent for accurate 3D geometry, and a radar agent for velocity-centric dynamics.

Table 2. Focus, core methods, and outputs of each perception agent. Each agent provides structured JSON facts $F_k$ and an optional synopsis $A_k$ .

| Agent | Focus & Core Methods | Outputs |
| --- | --- | --- |
| BEVFusion Agent | Global BEV geometric－semantic fusion; BEVFusion framework | $F_k$ : JSON for structured global scene elements; $A_k$ : natural-language summary of the global scene |
| Camera Perception Agent | Appearance & category semantics; object detection + scene-specific VLM (InsightGPT) | $F_k$ : JSON for detected objects; $A_k$ : semantic description and risk summary |
| LiDAR Perception Agent | Accurate 3D geometry & localization; 3D object detection on point clouds (CenterPoint) | $F_k$ : JSON for LiDAR obstacle detections; $A_k$ : summary of spatial layout |
| Radar Perception Agent | Radial velocity & hard-to-see dynamics; Density-based clustering (DBSCAN) | $F_k$ : JSON for radar detections; $A_k$ : description of motion trends of dynamic objects |

### 4.1.1 BEVFusion agent

The BEVFusion agent provides a global BEV-space representation by fusing multi-view camera features with LiDAR geometry. Following BEVFusion (Z. Liu, et al., 2023), it aligns heterogeneous sensor evidence in a common BEV/ego frame and outputs object-level hypotheses that are already geometrically consistent across the fused modalities. In addition, the agent produces a concise synopsis describing the global traffic layout, which is later used as auxiliary context for SSRE.

To remain consistent with our experimental focus, the BEVFusion agent primarily contributes object-centric facts (e.g., positions, sizes, headings, classes, and confidences). If additional static layout cues are available (e.g., coarse BEV semantics), they are treated as optional context rather than as mandatory outputs.

### 4.1.2 Camera perception agent

The camera agent focuses on appearance-centric semantics that are difficult to recover reliably from geometry alone, such as fine-grained categories and visual attributes. It first extracts image-space object hypotheses using an off-the-shelf detector, and then uses a traffic-scene vision-language model (Liu, et al., 2025) to generate a compact synopsis that highlights salient semantic cues (e.g., object type/attributes and risk-related observations).

For consistency with ICA, the camera agent reports structured facts in a unified schema (including local IDs, class probabilities/confidence, and optional depth/scale cues when available). The lifting from image space to the ego/BEV frame and the subsequent cross-sensor association are handled by ICA (Section 4.2), which avoids embedding non-deterministic geometry heuristics into the agent itself.

### 4.1.3 LiDAR perception agent

The LiDAR agent is responsible for accurate 3D geometry and localization. It applies a point-cloud 3D detector (CenterPoint (Yin, et al., 2021)) to produce precise ego-frame object hypotheses, including 3D centers, dimensions, headings, motion attributes (when available), and confidence scores. These structured facts provide high-quality geometric anchors for subsequent entity alignment and attribute fusion in ICA.

In parallel, the agent produces a short synopsis that summarizes the spatial layout and salient obstacles (e.g., closest leading vehicles or vulnerable road users), complementing the camera agent's appearance-centric cues.

### 4.1.4 Radar perception agent

The radar agent specializes in velocity-centric dynamics and provides redundancy under poor visibility (L. Wang, et al., 2026). It clusters radar returns using a lightweight density-based method (DBSCAN) and estimates per-cluster radial Doppler velocity together with a representative position. The resulting structured facts encode motion evidence that is often hard to obtain reliably from cameras alone and can complement LiDAR in adverse conditions.

Since radar measurements are naturally expressed in radar coordinates (range/azimuth/radial velocity), ICA performs the deterministic coordinate normalization into the ego/BEV frame before entity association and attribute fusion (Section 4.2.1). The radar synopsis emphasizes motion trends (e.g., approaching or receding targets) and is used as auxiliary context for SSRE when needed.

Overall, the multi-agent perception module provides complementary, typed evidence while keeping the interface compact and coordination-friendly. By separating structured facts (for deterministic coordination) from natural-language synopses (for auxiliary semantics), it supplies ICA and SSRE with consistent inputs and reduces the burden of reasoning over raw, redundant, and potentially contradictory perception outputs.

### 4.2 Information coordination and abstraction module

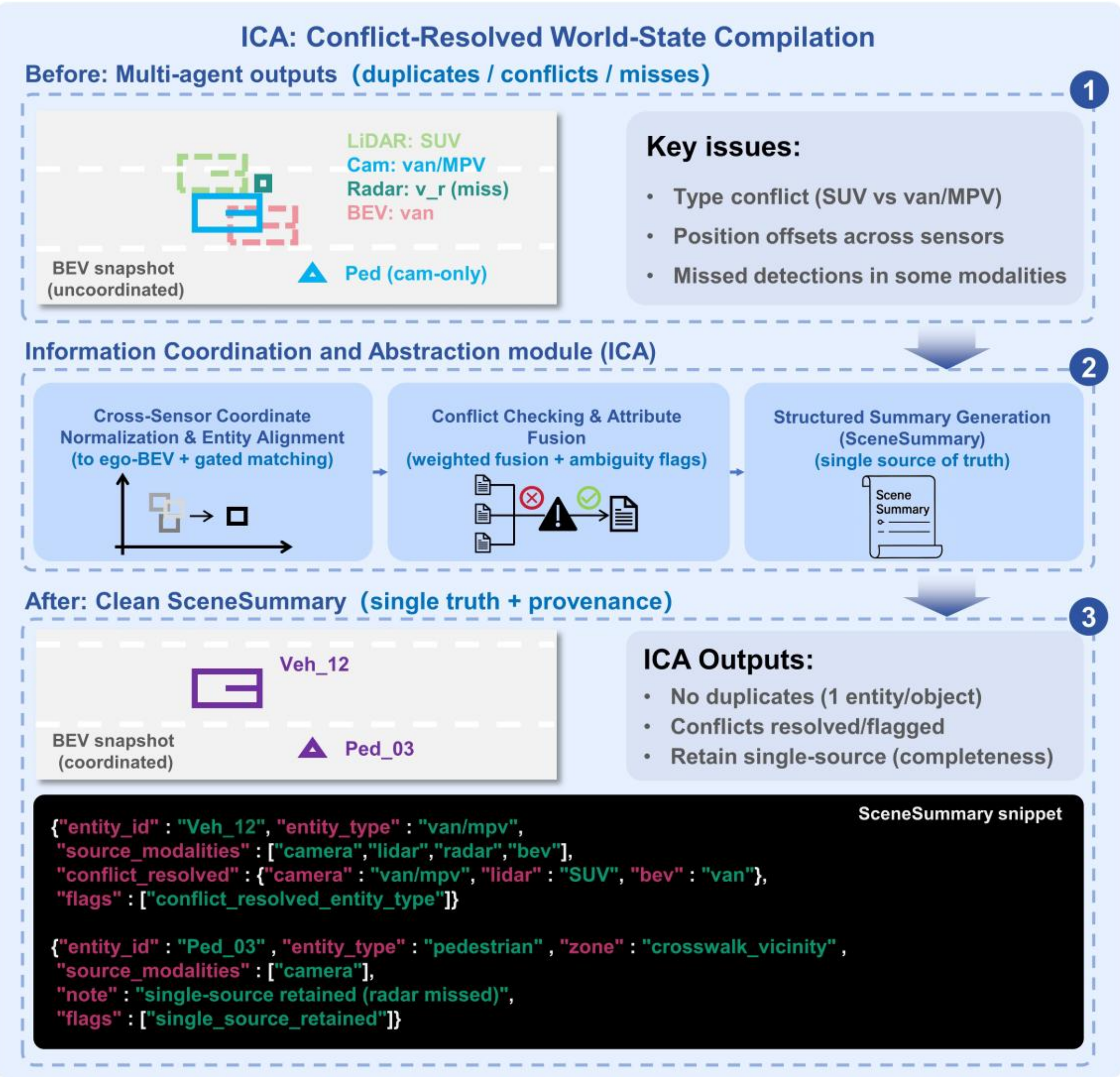

Figure 3. Information Coordination and Abstraction (ICA) pipeline: compiling uncoordinated multi-agent perception evidence into a conflict-resolved, provenance-aware SceneSummary.

The Information Coordination and Abstraction (ICA) module addresses semantic inconsistency that naturally arises when heterogeneous perception agents report overlapping, incomplete, or conflicting observations. Given the structured facts $\{F_k\}_{k=1}^{K}$ (JSON) produced by the multi-agent perception layer, ICA deterministically compiles them into a globally consistent world-state representation $S$ (SceneSummary) that serves as the single source of truth for downstream reasoning. To maximize reliability and auditability, ICA performs coordinate alignment and multi-source fusion using rule-based and uncertainty-aware operations, while reserving the large language model for high-level reasoning rather than low-level data reconciliation. Figure 3 illustrates an abstracted end-to-end view of ICA. Starting from uncoordinated multi-agent outputs (duplicates, conflicts, and misses), ICA sequentially performs (i) cross-sensor coordinate normalization and entity alignment, (ii) attribute-level fusion and conflict resolution, and (iii) deterministic structured-summary generation, yielding a compact and provenance-aware SceneSummary for subsequent SSRE reasoning.

**4.2.1 Cross-sensor coordinate normalization and entity alignment**

Since raw observations are reported in sensor-specific coordinate systems, ICA first maps all geometry-bearing detections (LiDAR/radar and BEVFusion outputs) into a unified ego-centric BEV reference. Camera detections remain in the image plane and are linked to the BEV world state through projection-based association, rather than depth-based back-projection. Let $\mathbf{p}_k$ be a point (or object center) in sensor frame $k$. Its coordinate in the ego frame is obtained by the calibrated extrinsics:

$$\mathbf{p}^{ego} = \mathbf{R}_{ego\leftarrow k}\,\mathbf{p}^{(k)} + \mathbf{t}_{ego\leftarrow k}. \tag{4}$$

For camera detections, ICA does not assume per-pixel or per-box depth availability. Instead, camera observations remain in the image plane. During Stage-3 association, ICA projects each 3D seed entity (from LiDAR/BEVFusion/radar-fused geometry) into the camera view using calibrated intrinsics K and extrinsics $\left(\mathbf{R}_{ego\leftarrow cam} + \mathbf{t}_{ego\leftarrow cam}\right)$:

$$\tilde{\mathbf{u}} = \pi\left(\mathbf{K}\left(\mathbf{R}_{ego\leftarrow cam}\,\mathbf{p}^{ego} + \mathbf{t}_{ego\leftarrow cam}\right)\right), \tag{5}$$

where $\pi\left(\left[x,y,z\right]^{\top}\right)=\left[x/z, y/z\right]^{\top}$ denotes perspective division. The projected 2D footprint $\tilde{b}_j$ of a 3D box is obtained by projecting its 3D corners and taking the axis-aligned bounding rectangle. Camera boxes $b_i^{cam}$ are matched to $\tilde{b}_j$ by IoU and class compatibility (Hungarian assignment). Matched camera observations contribute appearance-level semantics (e.g., color and fine-grained type) while the BEV geometry remains anchored to the 3D seeds.

For radar targets measured by range $r$, azimuth $\theta$ and radial velocity $v_{rad}$, ICA converts them to Cartesian coordinates, maps them to the ego frame, and approximates the BEV velocity direction by projecting the line-of-sight onto the BEV plane:

$$\mathbf{p}^{rad} = \left[r\cos\theta, r\sin\theta, 0\right]^{\top}, \quad \mathbf{p}^{ego} = \mathrm{R}_{ego\leftarrow rad}\,\mathbf{p}^{rad} + \mathbf{t}_{ego\leftarrow rad}, \quad \mathbf{v}^{bev} \approx v_{rad}\frac{\Pi(\mathbf{p}^{ego} - \mathbf{o}_{rad}^{bev})}{\left\|\Pi(\mathbf{p}^{ego} - \mathbf{o}_{rad}^{bev})\right\|_2}, \tag{6}$$

where $\Pi(\cdot)$ projects a 3D point onto the BEV plane and $\mathbf{o}_{rad}^{bev}$ denotes the radar origin in the ego-BEV coordinates. If the denominator is close to zero, ICA uses a minimum-norm fallback or skips the velocity vector to avoid numerical instability.

After normalization, ICA performs hierarchical entity alignment to merge detections that correspond to the same physical object. A spatial index (BEV grid or KD-tree) is built over ego-frame centers to prune candidate associations. For a detection $i$, we define its candidate neighborhood under a gating radius $r_g$ as:

$$\mathcal{N}(i) = \left\{ j \mid \left\|\mathbf{p}_i - \mathbf{p}_j\right\|_2 \le r_g \right\}. \tag{7}$$

The pairwise distance metric prioritizes Mahalanobis distance when a covariance estimate is available; otherwise Euclidean distance is used:

$$d(i,j) = \begin{cases} (\Delta\mathbf{p})^{\top}\,\Sigma^{-1}\,(\Delta\mathbf{p}), & covariance\ known, \\ \left\|(\Delta\mathbf{p})\right\|_2, & otherwise, \end{cases} \quad \Delta\mathbf{p} = \mathbf{p}_i - \mathbf{p}_j. \tag{8}$$

A composite matching cost further penalizes class mismatch and size disagreement under a distance gate $T_d$ :

$$C_{ij} = d(i,j) + \lambda_{cls}\left[c_i \neq c_j\right] + \lambda_{size}\left\|s_i - s_j\right\|_2, s.t.\ d(i,j) \leq T_d. \tag{9}$$

At each stage, ICA solves a linear assignment problem via the Hungarian algorithm:

$$\min_{\mathbf{X}} \sum_i \sum_j C_{ij} X_{ij} \quad s.t. \sum_j X_{ij} \leq 1,\ \sum_i X_{ij} \leq 1,\ X_{ij} \in \{0,1\}. \tag{10}$$

Unmatched LiDAR/radar/BEV detections are preserved as single-source BEV entities with explicit provenance tags. Unmatched 2D camera detections are retained as image-only semantic observations and are not exported as standalone BEV entities unless a geometry-bearing estimate (e.g., a camera-only 3D detector) is available.

Following the hierarchical strategy, ICA progressively fuses four agent streams: (1) LiDAR-BEVFusion matching establishes geometry-seed entities; (2) radar detections are attached to seeds with an additional velocity-consistency constraint; and (3) camera 2D detections are finally linked to BEV seeds by matching each 2D box to the 2D projection of a 3D seed (IoU-based cost with class-compatibility constraints), thereby enriching entities with appearance-level semantics without requiring camera depth. This staged design ensures that geometric consistency is established first, while modality-specific attributes are integrated only when association confidence is sufficient.

### 4.2.2 Conflict checking and attribute fusion

After entity association, each fused entity may contain multiple observations for the same attribute (e.g., position, velocity, class) from different sources. ICA resolves attribute-level conflicts by weighted evidence fusion, where the unnormalized weight of source $k$ for attribute $\alpha$ is:

$$\tilde{w}_k^{(\alpha)} = m_k^{(\alpha)} \cdot r_k \cdot q_k \cdot u_k^{(\alpha)} \cdot \eta_k^{(\alpha)}. \tag{11}$$

The normalized weight is computed as:

$$w_k^{(\alpha)} = \frac{\tilde{w}_k^{(\alpha)}}{\sum_{k'} \tilde{w}_{k'}^{(\alpha)}}. \tag{12}$$

Here $m_k^{(\alpha)}$ is a modality–attribute prior (e.g., LiDAR for geometry, radar for velocity, camera/BEVFusion for class), $r_k$ is a historical reliability score updated by an exponential moving average, $q_k$ denotes the current detection confidence, $u_k^{(\alpha)}$ quantifies uncertainty (from covariance when available, or a diagonal approximation from TTA/MC-Dropout), and $\eta_k^{(\alpha)}$ is a cross-modal consistency factor with respect to the current seed estimate.

The reliable term $r_k$ is updated as:

$$r_k^{(t)} = \beta r_k^{(t-1)} + (1-\beta) s_k^{(t)}, \tag{13}$$

where $s_k^{(t)} \in \{1, 0.5, 0\}$ denotes the pass/suspicious/veto score at frame $t$, computed by applying the residual gate in Section 4.2.1 with the current seed estimate $(\bar{\mathbf{x}}, \bar{\Sigma})$:

$$\eta_k^{(\alpha)} = \mathbf{exp}\left(-\left(\mathbf{x}_k^{(\alpha)} - \bar{\mathbf{x}}^{(\alpha)}\right)^\top \left(\bar{\Sigma}^{(\alpha)}\right)^{-1} \left(\mathbf{x}_k^{(\alpha)} - \bar{\mathbf{x}}^{(\alpha)}\right)\right) \tag{14}$$

For continuous attributes, ICA applies an information-weighted least-squares fusion:

$$\hat{\mathbf{x}}^{(\alpha)} = \left(\sum_k w_k^{(\alpha)} \left(\Sigma_k^{(\alpha)}\right)^{-1}\right)^{-1} \left(\sum_k w_k^{(\alpha)} \left(\Sigma_k^{(\alpha)}\right)^{-1} \mathbf{x}_k^{(\alpha)}\right), \tag{15}$$

with the fused covariance

$$\widehat{\Sigma}^{(\alpha)} = \left( \sum_k w_k^{(\alpha)} \left( \Sigma_k^{(\alpha)} \right)^{-1} \right)^{-1}. \tag{16}$$

To avoid over-confidence due to correlated sources (e.g., BEVFusion vs. its constituent LiDAR/camera streams), ICA uses Covariance Intersection (CI) when source lineage indicates correlation:

$$\left(\Sigma_{CI}\right)^{-1} = \omega\left(\Sigma_a\right)^{-1} + (1-\omega)\left(\Sigma_b\right)^{-1}, \ \mathbf{x}_{CI} = \Sigma_{CI}\left(\omega\left(\Sigma_a\right)^{-1}\mathbf{x}_a + (1-\omega)\left(\Sigma_b\right)^{-1}\mathbf{x}_b\right), \tag{17}$$

where $\omega \in (0,1)$ is chosen by a lightweight heuristic (defaulting to balanced fusion and optionally biased toward the lower-residual / higher-trust estimate).

For categorical attributes (e.g., object class), ICA uses weighted voting:

$$Conf(c) = \sum_k \omega_k^{(cls)} P_k(c), \ \widehat{c} = \mathbf{arg}\,\mathbf{max}_c\, Conf(c). \tag{18}$$

An ambiguity rule marks an attribute as uncertain when the class confidence is low or when any residual violates a $\chi^2$ gate under the fused covariance:

$$ambiguous \leftarrow \left(\mathbf{max}_c\, Conf(c) < \tau_{cls}\right) \vee \left(\left(\mathbf{x}_k - \widehat{\mathbf{x}}\right)^{\top} \widehat{\Sigma}^{-1}\left(\mathbf{x}_k - \widehat{\mathbf{x}}\right) > \chi^2_{d,\delta}\right). \tag{19}$$

### 4.2.3 Deterministic structured summary generation

Based on the fused entity set, ICA produces the final structured SceneSummary in a fully deterministic manner. Specifically, each entity is exported as a normalized record containing: a global entity ID, BEV position and velocity estimates (with uncertainty when available), class label and confidence, size, sensor provenance, fusion lineage, and ambiguity flags. The resulting SceneSummary is directly consumable by the central reasoning engine (SSRE) and enables transparent tracing from raw multi-agent evidence to the final coordinated world model.

Overall, the ICA module converts heterogeneous multi-agent perception outputs into a compact, globally consistent SceneSummary, providing a reliable factual substrate for downstream neuro-symbolic reasoning. A compact pseudocode of the full ICA pipeline is provided in Appendix A (Algorithm A.1).

### 4.3 SceneSummary-driven reasoning engine (SSRE)

The SceneSummary-Driven Reasoning Engine (SSRE) is designed to ensure that high-level driving reasoning is grounded on a consistent and auditable factual substrate. In contrast to conventional LLM-based pipelines that reason directly over raw, heterogeneous perception outputs, SSRE operates on a single, coordinated world-state representation $S$ (SceneSummary) produced by the Information Coordination and Abstraction (ICA) module (Section 4.2). This design explicitly decouples (i) uncertainty-prone multi-sensor reconciliation from (ii) open-domain semantic reasoning, thereby reducing the risk of hallucinated entities and untraceable justifications in safety-critical scenarios.

Existing LLM reasoning paradigms are effective at eliciting multi-step reasoning but typically assume that the input evidence is self-consistent. For example, Chain-of-Thought prompting encourages stepwise inference (Wei, et al., 2022), yet it remains vulnerable when the prompt contains ambiguous or contradictory claims. Similarly, agentic frameworks that interleave reasoning with actions (e.g., external tool calls) are not naturally aligned with autonomous driving, where trial-and-error interactions can be prohibitively costly (Yao, et al., 2022). Moreover, recent efforts toward interpretable driving decisions often provide reasoning traces but do not systematically verify whether each step is consistent with the perceived world state.

SSRE instantiates a symbolically grounded reasoning paradigm: it treats the coordinated SceneSummary $S$ as the single source of truth and enforces that all factual claims in the decision and justification are traceable to entities and attributes in $S$. Importantly, our coordination design does not rely on depth-based back-projection for camera observations; appearance semantics extracted from camera views are attached to BEV entities through projection-based association inside ICA, and SSRE only consumes the resulting BEV-anchored SceneSummary.

### 4.3.1 Interface specification

SSRE interacts with the rest of the system through a standardized interface. The inputs include: (i) an optional user query $Q$ in natural language, (ii) the coordinated SceneSummary $S$ generated by ICA, and (iii) an optional auxiliary perception synopsis $A$ (natural-language summaries from the perception agents). The outputs follow a dual-stream design:

(1) **StructuredDecision** $D_{final}$ (JSON), containing fields such as *recommended_action*, *confidence*, and *supporting_entity_ids*;

(2) **NaturalLanguageJustification** $N_{final}$, providing an audit-friendly explanation whose factual statements are explicitly anchored to entity IDs in $S$.

To preserve the "single-source-of-truth" principle, SSRE treats $A$ as non-binding context: it may improve readability or provide narrative cues, but it must not introduce new entities, numeric attributes, or relations that are absent from $S$. Any inconsistency between $A$ and $S$ is resolved in favor of $S$, and such inconsistencies are explicitly handled by the verification stage described below. All prompt specifications are provided in Appendix B.

Figure 4 illustrates the schema-level SSRE workflow. SSRE takes $(Q, S, A)$ as inputs and produces an entity-anchored decision $D_{final}$ and justification $N_{final}$ via a four-stage constrained prompting pipeline: Structured Scene Parsing (SSP), Systematic Risk Assessment (SRA), Verifiable Reasoning Chain Generation (VRCG), and Introspective Verification Loop (IVL).

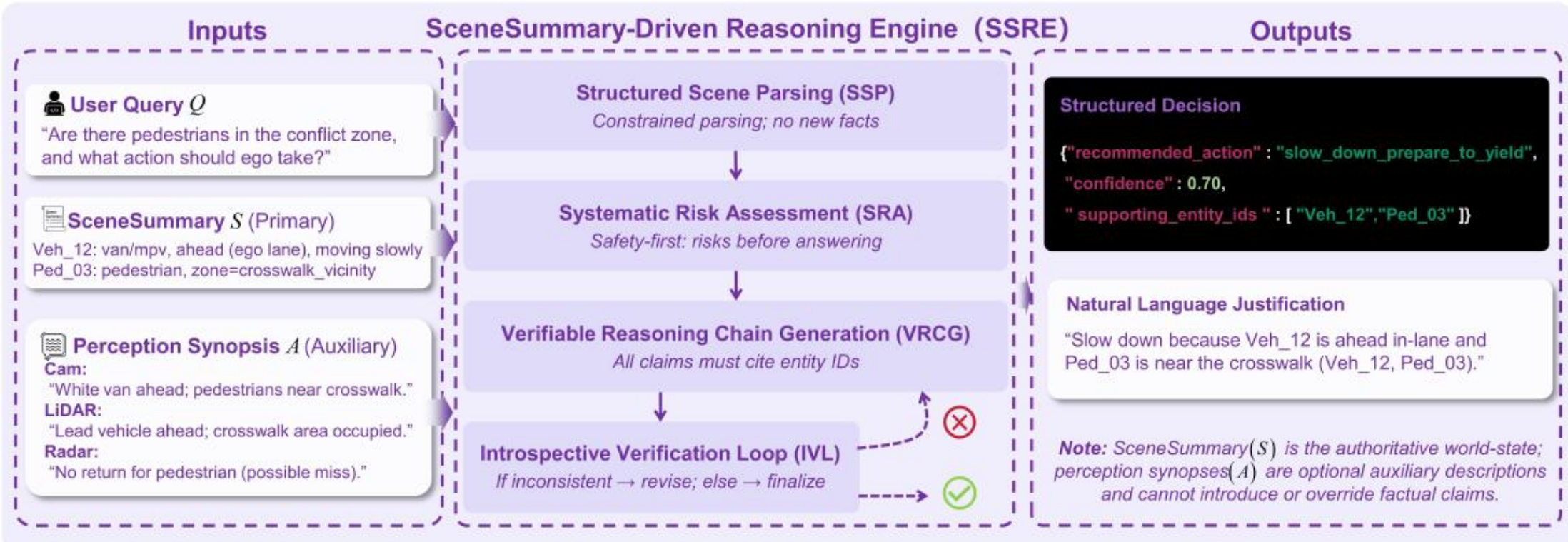

Figure 4. SceneSummary-Driven Reasoning Engine (SSRE): a four-stage verifier-aware reasoning workflow grounded on SceneSummary.

### 4.3.2 Four-stage verifiable reasoning workflow

SSRE follows a four-stage constrained prompting workflow that proceeds sequentially through Structured Scene Parsing (SSP), Systematic Risk Assessment (SRA), Verifiable Reasoning Chain Generation (VRCG), and the Introspective Verification Loop (IVL). All factual claims are anchored to SceneSummary entities and are iteratively verified before producing the final decision and justification (Algorithm 1).

---

Algorithm 1: Chain-of-Prompt Reasoning in SSRE

---

**Inputs:**

SceneSummary $S$ (JSON) *// coordinated world state (single source of truth)*

UserQuery $Q$ *// natural-language query/instruction*

AuxDescriptions $A$ (optional) *// auxiliary agent synopses (non-binding context)*

$k_max$ *// maximum IVL iterations*

$gamma = 0.5$ *// confidence penalty if not fully verified*

**Outputs:**

StructuredDecision $D_final$ *// JSON (recommended_action, confidence, supporting_entity_ids, ...)*

NaturalLanguageJustification $N_final$ *// final grounded justification text paired with D_final*

**Procedure:**

1: $M \leftarrow$ LLM.Generate( $T_parse$, $S$ )

*// SSP: constrained parsing of S into a standardized intermediate representation*

2: $L_risk \leftarrow$ LLM.Generate( $T_risk$, $M$ )

*// SRA: proactive risk identification and prioritization grounded in M*

3: $(D,N) \leftarrow$ LLM.Generate( $T_reason$, $(Q, M, L_risk, A)$ )

*// VRCG: draft decision + entity-anchored justification*

*// Requirement: every factual claim in N cites entity IDs from S / M*

4: verified ← false *// initialize the verification status for the IVL stage*

5: for k = 1 to k_max do *// run a bounded verify→revise loop to ensure termination*

6: $V \leftarrow$ LLM.Generate( $T_verify, (S,D,N)$ )

*//* $V$ *is a JSON object with fields:* $V.verdict \in \{Consistent, Minor, Major\}$, $V.comment$

7: if $V.verdict ==$ "Consistent" then

8: verified ← true

9: break

10: end if

11: $(D,N) \leftarrow$ LLM.Generate( $T_revision$, $(S, V, D, N)$ )

*// revise only the parts flagged by V; do not introduce new entities/attributes*

12: end for

13: $D_final \leftarrow D$ *// finalize the structured decision using the last revised draft*

14: $N_final \leftarrow N$ *// finalize the natural-language justification using the last revised draft*

15: if verified == false then *// handle the case where full consistency is not achieved within k_max*

16: $D_final$ .confidence ← gamma * D_final.confidence

17: $D_final$ .flags.add("unverified_after_k_max")

18: end if

19: return ( $D_final$ , $N_final$ ) *// output the final decision and its grounded justification*

---

### 4.3.3 Stage descriptions

**Structured Scene Parsing (SSP).** SSP uses a constrained template $T_parse$ to convert the SceneSummary JSON into a standardized intermediate representation $M$. Although $S$ is already structured, directly prompting an LLM with raw JSON may lead to field omissions or misinterpretations. SSP therefore normalizes the scene into a consistent format that explicitly lists entities, attributes, and relations, which reduces degrees of freedom in subsequent generation. In implementation, we set conservative decoding (e.g., low temperature) to improve stability.

**Systematic Risk Assessment (SRA).** Before answering the user query, SSRE forces an explicit risk assessment step using template $T_risk$. Based on $M$, the model enumerates potential hazards and outputs a ranked risk list $L_risk$ according to severity and urgency. This design injects safety considerations as a first-class constraint rather than an afterthought, aligning the reasoning process with safety-critical decision making.

**Verifiable Reasoning Chain Generation (VRCG).** VRCG is the core decision generation stage. Using template $T_reason$, SSRE produces a draft decision $D$ (JSON) and a draft justification $N$ in response to $Q$ while conditioning on $M$ and $L_risk$. Crucially, the prompt enforces entity anchoring: every factual statement about the scene must cite the corresponding entity ID, and the decision must include supporting_entity_ids that directly justify the recommended action. This constraint makes the reasoning traceable and prepares it for subsequent verification.

**Introspective Verification Loop (IVL).** IVL treats the LLM as a consistency verifier using template $T_verify$. Given $(S,D,N)$, the verifier checks whether any claim in the draft contradicts the ground-truth SceneSummary. If inconsistencies are found, SSRE triggers a revision step with $T_revision$ to update both the decision JSON and the justification while preserving all correct content. The loop runs for at most $k_{max}$ iterations. If full consistency is not achieved, SSRE still returns the latest draft but applies a confidence penalty and explicitly flags the output as not fully verified.

Overall, SSRE redefines the central LLM component as a SceneSummary-driven, verifier-aware reasoning engine. By grounding all reasoning on a coordinated world model and embedding risk-first and verification steps into the inference loop, SSRE provides traceable decisions and explanations that are substantially more robust to noisy evidence and misleading premises than unconstrained open-ended prompting.

## 5. Experimental results and analysis

This section describes the experimental setup used to evaluate InfoCoordiBridge, the proposed neuro-symbolic architecture, in which information coordination serves as the bridge between perception and reasoning. Specifically, we evaluate the complete pipeline from perception to reasoning, including multimodal 3D object detection, scene consistency analysis, open-domain question answering about driving scenes, and high level decision reasoning using the Information Coordination and Abstraction (ICA) module and the SceneSummary-Driven Reasoning Engine (SSRE).

### 5.1 Experimental setup

#### 5.1.1 Datasets

We conduct experiments on three public benchmarks—nuScenes (Caesar, et al., 2020), NuScenes-QA (Qian, et al., 2024), and the Waymo Open Dataset (Sun, et al., 2020)—to cover both multimodal perception and language-centric reasoning tasks. Specifically, nuScenes provides synchronized multi-sensor observations (6 surround cameras, 1 LiDAR, and 5 radars) with rich 3D annotations and diverse urban driving conditions, serving as the primary benchmark for evaluating multimodal perception performance. Furthermore, NuScenes-QA is built upon nuScenes 3D annotations and provides large-scale question–answer pairs (e.g., object attributes, spatial relations, risks, and decisions) together with scene semantic graphs; we treat the provided semantic graph as an oracle "ideal scene summary" to assess how well our generated SceneSummary matches the underlying scene semantics.

To examine cross-dataset robustness, we additionally use the Waymo Open Dataset, which contains dense annotations and a multi-sensor setup with high-resolution cameras and LiDAR (and notably differs from nuScenes by lacking radar), enabling an analysis of how sensor completeness affects fusion consistency. Following the conversion rules in Appendix C, we further derive a templated QA benchmark (Waymo-QA) whose task definitions and question styles are aligned with NuScenes-QA, so that generalization can be evaluated under comparable QA formats in a zero-shot setting.

#### 5.1.2 Baselines

For multimodal 3D detection and fusion consistency, we compare single-modality baselines (CenterPoint (Yin, et al., 2021) for LiDAR-only and RoPETR (Ji, et al., 2025) for camera-only), representative end-to-end camera–LiDAR fusion detectors (BEVFusion (Z. Liu, et al., 2023), TransFusion (Bai, et al., 2022), FSF (Fully Sparse Fusion) (Y. Li, et al., 2024), and GraphBEV(Song, et al., 2024)), and our multi-agent detector ensemble followed by ICA-based symbolic coordination. We also include an ablation variant without ICA to isolate the contribution of the information-coordination bridge to consistency improvements.

For cross-modal conflict resolution and scene-consistency evaluation, we include Naive Fusion and several representative perception-to-LLM pipelines (DriveVLM-style (Tian, et al., 2025), LLM4Drive-style (Yang, et al., 2023), and DriveGPT4-style (Xu, et al., 2024)). Notably, because public implementations of these systems are not fully compatible with our scene-consistency I/O constraints, we reproduce their core design ideas under a unified protocol and append the suffix "-style". Moreover, all baselines that require a text LLM are instantiated with the same open-source LLM backbones and evaluated in a zero-shot setting (without fine-tuning) to ensure a fair comparison. For open-ended QA and decision reasoning, we further compare pure VLM baselines (BLIP-2 (J. Li, et al., 2023), LLaVA (H. Liu, et al., 2023), Qwen2-VL (P. Wang, et al., 2024)), a BEV-based VLM (Talk2BEV-style (Choudhary, et al., 2024)), and planning-style agents (DriveGPT4-style (Xu, et al., 2024) and PlanAgent-style (Zheng, et al., 2024)), together with controlled ablations that toggle ICA, SSRE, and the Introspective Verification Loop (IVL).

### 5.1.3 Metrics

We adopt a multi-dimensional metric suite spanning accuracy, consistency, safety, reliability, and efficiency. For detection accuracy, we report mean Average Precision (mAP) and nuScenes Detection Score (NDS) on nuScenes, and mAP on Waymo. In addition, to quantify fusion consistency, we use Entity Redundancy Rate (ERR; lower is better) and Attribute Consistency Rate (ACR; higher is better), where ERR measures how often a physical object is redundantly reported after fusion, and ACR measures cross-sensor agreement on key attributes such as class, position, and heading.

For conflict-resolution benchmarking, we report Conflict Resolution Rate (CRR), Missed Detection Compensation Rate (MDCR), the average number of hallucinated entities per scene (HE; lower is better), and entity-level Precision/Recall/F1 (EP/ER/EF1). For unified-interface analysis, we measure answer Accuracy, Hallucination Rate (HR), and inference efficiency using Avg. Prompt Length (APL) and Avg. Inference Time (AIT).

Finally, for open-ended QA and high-level decision reasoning, we evaluate: (i) task-level correctness via QA Exact Match (QA EM), Answer F1, and Decision Accuracy (DA); (ii) explanation quality via BLEU-4 and ROUGE-L; (iii) safety via Risk Identification F1 (RI F1); and (iv) reliability & hallucination via Fact Consistency Pass Rate (FCPR) and Hallucinated Entity Mention Rate (HER; lower is better), which assess factual alignment of answers/decisions with the SceneSummary and the tendency to mention non-existent entities, respectively.

## 5.2 Multimodal 3D detection performance and fusion consistency evaluation

This experiment evaluates whether the proposed Information Coordination and Abstraction (ICA) module can fuse multi-sensor 3D detections into a globally consistent BEV world representation while maintaining competitive detection accuracy. We compare our "independent multi-agent detectors with ICA" pipeline against representative single-modality detectors and end-to-end camera–LiDAR fusion baselines under a unified BEV/ego coordinate system.

We conduct the evaluation on nuScenes as the primary benchmark and additionally test on the Waymo Open Dataset to examine cross-dataset robustness under different sensor configurations. Notably, nuScenes contains camera, LiDAR, and radar modalities, whereas Waymo contains camera and LiDAR only; this modality-completeness difference helps analyze how explicit coordination behaves when the number and heterogeneity of sensor sources changes.

As baselines, we include CenterPoint (LiDAR-only) and RoPETR (camera-only) to characterize single-modality performance limits, and we further compare several representative end-to-end camera–LiDAR fusion detectors, including TransFusion, BEVFusion, FSF, and GraphBEV. For our approach, we report two variants: Ours (without ICA), which aggregates per-agent detections without explicit cross-sensor coordination, and Ours, which applies ICA to perform coordinate normalization, entity association, and attribute fusion before producing the final fused detection list.

We report both detection accuracy metrics and fusion-consistency metrics. For detection accuracy, we use the official nuScenes metrics mean Average Precision (mAP) and nuScenes Detection Score (NDS), and we use mAP on Waymo. For consistency, we use Entity Redundancy Rate (ERR; lower is better), which measures the fraction of duplicated reports for the same physical object in the final detection list, and Attribute Consistency Rate (ACR; higher is better), which measures cross-sensor agreement on key attributes such as class, position, and heading.

Therefore, higher mAP/NDS/ ACR and lower ERR indicate better overall performance.

Table 3. Performance and consistency of multimodal 3D detection methods on the nuScenes and Waymo datasets. Best results are shown in bold, and second-best results are shown in italics. "L" denotes LiDAR and "C" denotes camera; "–" indicates that ERR/ACR are not applicable for single-modality methods.

| | nuScenes | | | | Waymo | | |
|---|---|---|---|---|---|---|---|
| Method | mAP(%) | NDS(%) | ERR(%) | ACR(%) | mAP(%) | ERR(%) | ACR(%) |
| CenterPoint (L-only) | 60.3 | 67.3 | - | - | 68.4 | - | - |
| RoPETR (C-only) | 64.8 | 70.9 | - | - | 69.7 | - | - |
| TransFusion (L+C) | 68.9 | 71.7 | 18.0 | 82.0 | 65.1 | 10.0 | 85.0 |
| BEVFusion (L+C) | 70.2 | 72.9 | 15.0 | 85.0 | **70.0** | 8.0 | 90.0 |
| FSF (L+C) | 70.4 | 72.7 | 12.0 | 88.0 | **70.0** | 5.0 | 92.0 |
| GraphBEV (L+C) | **71.7** | **73.6** | 10.0 | 90.0 | **70.0** | 5.0 | 90.0 |
| Ours (without ICA) | 58.0 | 63.0 | 25.0 | 75.0 | 67.0 | 30.0 | 70.0 |
| **Ours** | 70.9 | 73.2 | **0.7** | **98.0** | 69.8 | **1.0** | **95.0** |

On nuScenes, GraphBEV achieves the highest detection accuracy (mAP=71.7%, NDS=73.6%). Our method reaches mAP=70.9% and NDS=73.2%, which is within 1 percentage point of the strongest end-to-end BEV baselines, indicating that introducing ICA does not compromise 3D detection accuracy. On Waymo, several end-to-end fusion baselines reach mAP=70.0%; our method attains mAP=69.8%, remaining competitive under cross-dataset transfer.

The consistency metrics highlight the key advantage of symbolic coordination. Without ICA, simply aggregating the multi-agent detections yields substantial redundancy and conflicts: on nuScenes, Ours (without ICA) exhibits ERR=25.0% and ACR=75.0%; on Waymo, ERR further increases to 30.0% and ACR drops to 70.0%, indicating that naïve union of detections produces many duplicated boxes and inconsistent attributes. End-to-end fusion baselines partially mitigate this issue through feature-level fusion, but still retain non-negligible redundancy (ERR≈10–18% on nuScenes and 5–10% on Waymo) and limited attribute agreement (ACR≈82–90% on nuScenes and 85–92% on Waymo). In contrast, our ICA-based method reduces ERR to 0.7% and increases ACR to 98.0% on nuScenes; it also achieves ERR=1.0% and ACR=95.0% on Waymo, outperforming all compared approaches on consistency across both benchmarks.

Notably, comparing nuScenes and Waymo suggests that modality completeness amplifies the benefit of explicit coordination. In nuScenes, the presence of additional sensor sources (including radar) increases the chance of cross-source redundancy and attribute disagreement, and ICA yields the most pronounced gains on ERR/ACR. In Waymo, where the sensor set is smaller (camera with LiDAR), the end-to-end fusion baselines already show improved consistency; nevertheless, ICA still further compresses redundancy and improves attribute agreement, indicating that the coordination mechanism remains beneficial even when the modality set is reduced.

Overall, this experiment shows that the ICA module functions as an effective "information coordination bridge" between multimodal perception and downstream reasoning: it maintains detection accuracy at a level comparable to strong end-to-end fusion baselines while producing a substantially more compact and self-consistent fused world representation (low ERR and high ACR). This consistency gain is essential for the subsequent SceneSummary-driven neuro-symbolic reasoning, where redundant or conflicting entities can directly trigger unreliable high-level inferences.

### 5.3 Cross-modal conflict resolution and scene consistency analysis

This experiment moves beyond the overall detection accuracy reported in Section 5.2 and focuses on semantic reliability of multi-sensor fusion. Specifically, we evaluate whether fused perception outputs form a self-consistent, contradiction-free scene description, which is essential when the downstream reasoning module relies on a single "source of truth" (SceneSummary) rather than reconciling noisy evidence on its own.

We instantiate this evaluation as a scene fact-consistency verification task on NuScenes-QA. For each driving scene, a method produces a fused entity set and a scene-level representation. We then quantify how effectively cross-modal conflicts are resolved, how many missed entities are recovered, and how often hallucinated entities are introduced when LLMs are used in the pipeline.

We compare several representative fusion and perception-to-LLM paradigms: Naive Fusion (direct union of detections without explicit coordination), Naive Fusion + LLM (direct prompting baseline that asks a text LLM to consolidate the naïvely aggregated multi-source detections into a single entity list under the same output schema), DriveVLM-style, LLM4Drive-style, DriveGPT4-style, and our method (Ours). Furthermore, all baselines that require a text LLM use the same open-source LLM backbones and are evaluated in a strict zero-shot setting to ensure fair comparison.

Table 4 reports Conflict Resolution Rate (CRR, %), Missed Detection Compensation Rate (MDCR, %), the average number of Hallucinated Entities per scene (HE, lower is better), and entity-level Precision/Recall/F1 (EP/ER/EF1, %). Here, CRR and MDCR are computed with respect to an oracle entity set $G$ from the provided NuScenes-QA scene graph, using the same one-to-one entity matching operator $\pi(\cdot)$ as EP/ER/EF1. Specifically, $\pi$ performs class-consistent BEV IoU matching with Hungarian assignment under an IoU threshold $\tau$ . For each scene $s$ , let $C^{(s)} \subseteq G^{(s)}$ denote conflict entities for which at least two input sources yield reports matched to the same oracle entity but disagree on key attributes (e.g., class or heading beyond a tolerance), and let $C_{fix}^{(s)} \subseteq C^{(s)}$ denote those whose final fused output yields a single matched entity whose key attributes are oracle-consistent. We report Conflict Resolution Rate as a micro-average over the evaluation set:

$$CRR\ (\%) = \frac{\sum_s \left|C_{fix}^{(s)}\right|}{\sum_s \left|C^{(s)}\right|} \times 100. \tag{20}$$

Similarly, let $M^{(s)} \subseteq G^{(s)}$ denote oracle entities missed by the designated primary detector (the BEVFusion agent) under $\pi$ , and let $M_{comp}^{(s)} \subseteq M^{(s)}$ denote those recovered by the final fused output. We report Missed Detection Compensation Rate as:

$$MDCR\ (\%) = \frac{\sum_s \left|M_{comp}^{(s)}\right|}{\sum_s \left|M^{(s)}\right|} \times 100. \tag{21}$$

EP/ER/EF1 summarize the overall entity quality after fusion, while HE directly captures hallucination risks in perception-to-LLM pipelines. Higher is better for CRR, MDCR, EP, ER, and EF1; lower is better for HE.

Table 4. Cross-Modal Conflict Resolution and Scene Consistency Comparison on NuScenes-QA. Methods with the suffix "-style" are protocol-aligned reproductions under unified I/O constraints. "–" indicates not applicable for Naive Fusion (non-generative union), which does not synthesize entities.

| Method | CRR(%) | MDCR(%) | HE(per-scene) | EP(%) | ER(%) | EF1(%) |
|---|---|---|---|---|---|---|
| Naive Fusion | 0.0 | 36.5 | - | 78.2 | 92.5 | 84.7 |
| Naive Fusion + LLM | 2.3 | 37.8 | 0.9 | 74.5 | 79.1 | 76.7 |
| DriveVLM-style | 1.7 | 40.5 | 1.2 | 76.8 | 80.3 | 78.5 |
| LLM4Drive-style | 3.4 | 38.2 | 1.4 | 72.2 | 77.6 | 74.8 |
| DriveGPT4-style | 4.1 | 39.0 | 1.6 | 75.9 | 82.1 | 78.9 |
| Ours | **91.7** | **74.6** | **0.1** | **87.3** | **89.2** | **88.2** |

Table 4 shows a clear separation between explicit symbolic coordination and non-coordinated baselines. Without an explicit coordination stage, all compared baselines achieve CRR below 5%, indicating that most cross-modal contradictions remain unresolved and are directly passed to the final scene representation. In contrast, our method reaches CRR=91.7%, demonstrating that the ICA-based coordination effectively reconciles duplicated entities and conflicting attributes into a single consistent world model.

Furthermore, our method substantially improves completeness without introducing spurious entities. MDCR increases to 74.6%, nearly doubling the range of the baselines (36.5%–40.5%), which indicates that the coordinated fusion retains complementary detections and recovers missed entities more effectively. At the same time, our method produces almost no hallucinated entities (HE=0.1 per scene), whereas perception-to-LLM pipelines without a clean, coordinated world model generate 0.9–1.6 hallucinated entities per scene on average. This result supports the central motivation of our architecture: providing a clean, self-consistent SceneSummary prior to reasoning is critical to suppress hallucination-prone high-level inference.

This large margin is expected because CRR and MDCR are coordination-focused metrics rather than detection-accuracy metrics: they explicitly test whether a pipeline can (i) identify that multiple reports refer to the same physical entity, and (ii) reconcile mutually inconsistent attributes into a single, schema-compliant representation. Non-coordinated baselines either keep the raw union of multi-source detections (thus preserving contradictions by construction) or rely on an LLM to consolidate lists without deterministic geometric association; in practice, such consolidation is brittle under dense scenes and noisy cross-sensor discrepancies, often leaving conflicts unaligned or discarding complementary evidence. In contrast, ICA performs explicit coordinate normalization and entity alignment before any language reasoning, and fuses attributes under fixed rules with provenance/uncertainty bookkeeping, which directly targets the failure mode measured by CRR/MDCR and prevents contradictions from being propagated downstream.

Entity-level precision/recall further explain the reliability gap across paradigms. Naive Fusion obtains the highest recall (ER=92.5%) by construction because it largely preserves single-source detections; however, its precision is limited (EP=78.2%) due to duplicated boxes and unresolved attribute conflicts, which lowers the overall EF1. Perception-to-LLM pipelines slightly reduce redundancy but remain vulnerable to noisy and contradictory inputs, leading to both lower EP/ER and non-negligible hallucinations. In contrast, our method achieves a substantially better balance (EP=87.3%, ER=89.2%, EF1=88.2%) by explicitly aligning entities and fusing attributes during the coordination stage, which removes contradictory duplicates while preserving most valid detections.

Finally, we provide qualitative visualizations in Section 5.6.1 to illustrate how cross-modal conflicts and missed detections are handled in representative nuScenes scenes, and how these corrections translate into improvements in CRR/MDCR/HE (as well as the detection-level consistency metrics ERR/ACR in Section 5.2). Overall, this experiment demonstrates that inserting an explicit Information Coordination and Abstraction module between perception and reasoning yields a more reliable, auditable, and self-consistent scene representation, which forms an indispensable foundation for subsequent high-level reasoning and decision making.

### 5.4 Impact of the multi-agent unified interface on LLM reasoning performance

This experiment evaluates whether a unified, structured interface between multi-agent perception and large language model (LLM) reasoning improves open-domain driving question answering under a zero-shot setting.

The unified interface is instantiated as the output of our Information Coordination and Abstraction (ICA) module: a single, globally consistent SceneSummary represented in JSON, together with a concise natural-language scene synopsis aggregated from the perception agents. Both components are injected into the LLM prompt as plain text, enabling downstream reasoning without exposing the model to raw, redundant, or mutually inconsistent perception logs.

We adopt NuScenes-QA as the benchmark and compare (i) end-to-end vision–language models (VLMs) that answer from camera images (BLIP-2, LLaVA, Qwen2-VL), and (ii) text-only LLMs (LLaMA2-13B, Qwen-14B, GLM-4-9B, DeepSeek-14B). For each text-only LLM, we evaluate two prompting conditions: (a) no unified interface, where the prompt contains concatenated raw multi-agent detection text (uncoordinated and potentially redundant/conflicting), and (b) with unified interface, where the prompt contains the SceneSummary JSON, the scene synopsis, and the question. All models are evaluated strictly in a zero-shot manner without fine-tuning on NuScenes-QA.

We report Accuracy and F1 for task correctness, Hallucination Rate (HR) for reliability, and Average Prompt Length (APL) / Average Inference Time (AIT) for efficiency. HR measures the fraction of answers that mention entities or relations not supported by the scene. APL counts the number of text tokens in the prompt and is therefore not applicable to image-conditioned VLM baselines.

Table 5. Performance of different methods on the NuScenes-QA subset under the zero-shot setting.

| Method | Accuracy (%) | F1 | HR (%) | APL (tokens) | AIT (s) |
|---|---|---|---|---|---|
| BLIP-2 | 30 | 0.40 | 15 | - | 4.7 |
| LLaVA | 35 | 0.45 | 20 | - | 3.5 |
| Qwen2-VL | 40 | 0.50 | 15 | - | 4.0 |
| LLaMA2-13B (no unified interface) | 45 | 0.50 | 30 | ~1000 | 6.5 |
| LLaMA2-13B (with unified interface) | 65 | 0.70 | 5 | ~260 | 4.0 |
| Qwen-14B (no unified interface) | 50 | 0.55 | 25 | ~1000 | 7.0 |
| Qwen-14B (with unified interface) | 75 | 0.80 | 5 | ~260 | 4.5 |
| GLM-4-9B (no unified interface) | 30 | 0.35 | 35 | ~1000 | 4.0 |
| GLM-4-9B (with unified interface) | 55 | 0.60 | 10 | ~260 | 2.5 |
| DeepSeek-14B (no unified interface) | 55 | 0.60 | 20 | ~1000 | 6.0 |
| DeepSeek-14B (with unified interface) | 78 | 0.85 | 5 | ~260 | 3.5 |

Table 5 summarizes the zero-shot results on NuScenes-QA. Notably, the unified interface consistently improves text-only LLM reasoning across all backbones. Relative to raw detection text prompting, Accuracy increases by 20–25% (e.g., from 45% to 65% for LLaMA2-13B and from 50% to 75% for Qwen-14B), while HR drops sharply to 5–10% (e.g., 30% to 5% and 35% to 10%). Furthermore, GLM-4-9B with the unified interface (55% Accuracy) surpasses Qwen-14B without it (50% Accuracy), indicating that input quality and structure can partially compensate for smaller model capacity.

In contrast, end-to-end VLM baselines remain substantially weaker on this benchmark (30–40% Accuracy with 15–20% HR), which is consistent with their limited ability to perform structured 3D/spatial reasoning and multi-view evidence consolidation in complex driving scenes when an explicit, typed world state is absent.

The efficiency results further support the benefit of the interface. By replacing long, uncoordinated detection logs (~1000 tokens) with a compact SceneSummary-plus-synopsis prompt (~260 tokens), the unified interface reduces APL by approximately 74% (~3.8× shorter) and yields a 36–42% reduction in inference latency (AIT) across the evaluated LLMs.

Overall, these results show that a unified SceneSummary interface functions as an effective "information bridge" for LLM-based driving cognition: it improves answer correctness, suppresses hallucinations by removing redundancy and cross-modal inconsistency at the interface, and reduces computational overhead through prompt compression.

### 5.5 Open-world driving scene QA and high-level decision reasoning performance

This experiment evaluates whether the proposed neuro-symbolic architecture supports reliable open-world driving question answering (QA) and safety-critical high-level decision reasoning. Specifically, we assess how the Information Coordination and Abstraction (ICA) module and the SceneSummary-Driven Reasoning Engine (SSRE) jointly improve task correctness, safety awareness, and factual reliability under open-ended natural-language queries.

Given a scene representation and a user query, the system outputs (i) an answer or a recommended action and (ii) a traceable natural-language justification. We consider four question types: (1) Relation QA, which queries spatial relations among entities; (2) Dynamic/Status QA, which queries motion states and temporal dynamics; (3) Risk Prediction QA, which asks for potential hazards and risk explanations; and (4) Decision QA, which requires a high-level driving suggestion under traffic-rule constraints (e.g., whether to slow down or stop).

We mainly evaluate on NuScenes-QA. Furthermore, to test cross-dataset robustness, we convert the Waymo Open Dataset into a NuScenes-QA-style benchmark (Waymo-QA) using aligned templates (see the conversion templates in the appendix C). This design keeps task definitions consistent across datasets, enabling a controlled assessment of both in-domain performance and transfer robustness.

Table 6 summarizes the compared methods. We include (A1–A3) pure vision-language models (BLIP-2, LLaVA, Qwen2-VL) that answer directly from multi-view images; (A4) a BEV-based VLM baseline (Talk2BEV-style); (B1–B2) multimodal planning-style agents (DriveGPT4-style and PlanAgent-style); and (C1–C5) our variants that isolate the contribution of each architectural component. In particular, C2 applies ICA to produce a clean SceneSummary followed by a single-stage chain-of-thought, C3 runs the full SSRE workflow on a noisy pseudo-SceneSummary without ICA, C5 removes the Introspective Verification Loop (IVL), and C4 is the full ICA+SSRE system. For fairness, all methods requiring a text LLM are evaluated in a zero-fine-tuning setting using the same open-source LLM family.

Table 6. Compared methods in Experiment 5.5 (w/o = without).

| ID | Method | Paradigm / Type | Brief description |
|---|---|---|---|
| A1 | BLIP-2 | Pure VLM baseline | Pre-trained vision–language model answering questions directly from multi-view camera images and text. |
| A2 | LLaVA | Pure VLM baseline | Instruction-tuned VLM performing open-ended VQA on images without explicit BEV or 3D structure. |
| A3 | Qwen2-VL | Pure VLM baseline | Bilingual VLM used as a generic VQA backbone on camera images and text questions. |
| A4 | Talk2BEV-style | BEV-based VLM | BEV-based VQA baseline that feeds rendered BEV semantic maps and questions into a general VLM. |
| B1 | DriveGPT4-style | End-to-end V-L driving agent | End-to-end vision–language driving agent generating scene descriptions and high-level driving suggestions. |
| B2 | PlanAgent-style | Structured planning agent | Planning agent using structured environment descriptions with hierarchical chain-of-thought and a single reflection step. |
| C1 | Ours (Raw-Text) | LLM agent w/o ICA or SSRE | LLM agent reasoning on concatenated raw detection logs and textual summaries without information coordination or SSRE. |
| C2 | Ours (w/o SSRE) | ICA-only ablation | Applies the ICA module to build a clean SceneSummary, then runs a single-stage chain-of-thought. |
| C3 | Ours (w/o ICA) | SSRE-only ablation | Uses the full SSRE workflow on a noisy pseudo-SceneSummary built from uncoordinated detections. |
| C4 | Ours | Full ICA + SSRE | Full method combining ICA with the four-stage SSRE workflow for verifiable reasoning and decision making. |
| C5 | Ours (w/o IVL) | Ablation of IVL | Ablated version of Ours that removes the Introspective Verification Loop from SSRE. (For ablation study only) |

We report four families of metrics. (i) Task-level correctness uses QA EM (Exact Match) and Answer F1 for QA tasks, and Decision Accuracy (DA) for Decision QA. (ii) Explanation similarity uses BLEU-4 and ROUGE-L to quantify overlap with reference justifications. (iii) Safety is measured by Risk Identification F1 (RI F1), which evaluates risk detection performance. (iv) Reliability and hallucination are measured by Fact Consistency Pass Rate (FCPR) and Hallucinated Entity Mention Rate (HER; lower is better), which explicitly assess whether the decision and explanation remain consistent with the SceneSummary and whether the model mentions entities not present in the scene representation.

### 5.5.1 Overall performance on NuScenes-QA and Waymo-QA

Table 7. Overall performance of compared methods on the NuScenes-QA dataset (%).

| ID | Method | QA EM | Answer F1 | DA | BLEU-4 | ROUGE-L | RI F1 | FCPR | HER |
|---|---|---|---|---|---|---|---|---|---|
| A1 | BLIP-2 | 42.3 | 60.8 | 50.4 | 20.2 | 46.1 | 38.5 | 58.2 | 36.7 |
| A2 | LLaVA | 46.1 | 63.4 | 52.1 | 21.8 | 47.9 | 41.2 | 61.0 | 34.6 |
| A3 | Qwen2-VL | 47.8 | 64.9 | 54.3 | 23.1 | 49.2 | 42.0 | 62.8 | 33.9 |
| A4 | Talk2BEV-style | 56.4 | 70.6 | 60.8 | 26.4 | 53.5 | 54.7 | 74.3 | 20.8 |
| B1 | DriveGPT4-style | 52.6 | 67.3 | 69.4 | **31.2** | **58.4** | 58.6 | 76.1 | 21.5 |
| B2 | PlanAgent-style | 55.7 | 70.2 | 72.6 | 28.4 | 55.6 | 63.5 | 82.4 | 46.9 |
| C1 | Ours (Raw-Text) | 49.1 | 66.2 | 60.3 | 27.1 | 52.0 | 50.6 | 55.1 | 41.8 |
| C2 | Ours (w/o SSRE) | 57.3 | 72.4 | 68.2 | 26.2 | 53.2 | 57.4 | 79.3 | 17.6 |
| C3 | Ours (w/o ICA) | 58.1 | 73.1 | 70.4 | 26.8 | 54.0 | 59.2 | 81.2 | 15.9 |
| C4 | Ours | **64.3** | **77.5** | **76.2** | <u>28.9</u> | <u>56.3</u> | **66.4** | **93.2** | **7.4** |

Table 8. Overall performance of compared methods on the Waymo-QA dataset (%).

| ID | Method | QA EM | Answer F1 | DA | BLEU-4 | ROUGE-L | RI F1 | FCPR | HER |
|---|---|---|---|---|---|---|---|---|---|
| A1 | BLIP-2 | 39.2 | 57.6 | 46.4 | 19.1 | 44.0 | 35.7 | 55.3 | 38.9 |
| A2 | LLaVA | 42.1 | 60.2 | 48.5 | 20.7 | 45.8 | 38.4 | 57.9 | 36.8 |
| A3 | Qwen2-VL | 44.3 | 61.9 | 50.1 | 21.8 | 46.7 | 39.2 | 59.0 | 35.6 |
| A4 | Talk2BEV-style | 53.2 | 67.1 | 56.3 | 24.6 | 51.2 | 50.7 | 71.3 | 22.7 |
| B1 | DriveGPT4-style | 49.4 | 65.2 | 66.5 | **30.4** | **55.6** | 55.4 | 73.2 | 23.4 |
| B2 | PlanAgent-style | 52.5 | 67.4 | 70.2 | 27.2 | 53.4 | 60.6 | 79.4 | 18.6 |
| C1 | Ours (Raw-Text) | 46.1 | 63.1 | 56.0 | 24.8 | 49.7 | 47.2 | 52.4 | 43.7 |
| C2 | Ours (w/o SSRE) | 54.2 | 69.3 | 65.4 | 25.0 | 51.0 | 53.5 | 76.1 | 19.5 |
| C3 | Ours (w/o ICA) | 55.1 | 70.1 | 67.2 | 25.4 | 52.0 | 55.2 | 78.3 | 17.9 |
| C4 | Ours | **61.3** | **75.2** | **73.1** | 27.8 | 54.1 | **62.4** | **91.4** | **8.6** |

Table 7 and Table 8 report the overall results. On NuScenes-QA, the full system (C4) achieves the best performance across correctness, safety, and reliability, reaching 64.3% QA EM, 77.5% Answer F1, 76.2% DA, and 66.4% RI F1, while also obtaining the highest FCPR (93.2%) and the lowest HER (7.4%). Notably, the strongest non-ours baseline (PlanAgent-style) remains behind C4 on QA and reliability (55.7% QA EM, 82.4% FCPR, and 46.9% HER), indicating that planning-oriented reasoning alone does not guarantee factual grounding under open-ended queries.

On Waymo-QA, the same trend holds. C4 remains the best performer, achieving 61.3% QA EM, 75.2% Answer F1, 73.1% DA, and 62.4% RI F1, together with strong reliability (91.4% FCPR) and low hallucination (8.6% HER). In contrast, PlanAgent-style attains relatively high DA (70.2%) but shows weaker overall QA and reliability compared to C4 (52.5% QA EM and 79.4% FCPR). These results suggest that combining a clean, coordinated world model with verifier-aware reasoning provides more dependable open-world QA and decision making under domain shift.

### 5.5.2 Task-type breakdown

Table 9. Task-type-wise performance of compared methods on NuScenes-QA (%).

| | | Relation QA | | Dynamic/Status QA | | Risk Prediction QA | | | Decision QA |
|---|---|---|---|---|---|---|---|---|---|
| ID | Method | EM | F1 | EM | F1 | EM | F1 | RI F1 | Accuracy |
| A4 | Talk2BEV-style | 60.8 | 74.6 | 55.3 | 70.4 | 48.1 | 66.3 | 54.7 | 60.8 |
| B2 | PlanAgent-style | 58.6 | 72.0 | 54.1 | 68.8 | 54.0 | 70.2 | 63.5 | 72.6 |
| C1 | Ours (Raw-Text) | 50.2 | 67.3 | 46.1 | 63.8 | 40.4 | 60.1 | 50.6 | 60.3 |
| C2 | Ours (w/o SSRE) | 61.2 | 75.0 | 58.4 | 73.1 | 50.6 | 68.0 | 57.4 | 68.2 |
| C3 | Ours (w/o ICA) | 59.4 | 74.3 | 57.6 | 72.4 | 52.1 | 69.1 | 59.2 | 70.4 |
| C4 | Ours | **67.3** | **80.6** | **64.2** | **78.1** | **60.5** | **74.0** | **66.4** | **76.2** |

Table 10. Task-type-wise performance of compared methods on Waymo-QA (%).

| | | Relation QA | | Dynamic / Status QA | | Risk Prediction QA | | | Decision QA |
|---|---|---|---|---|---|---|---|---|---|
| ID | Method | EM | F1 | EM | F1 | EM | F1 | RI F1 | Accuracy |
| A4 | Talk2BEV-style | 57.1 | 71.2 | 52.0 | 67.1 | 45.0 | 63.0 | 50.7 | 56.3 |
| B2 | PlanAgent-style | 55.3 | 69.0 | 51.0 | 66.2 | 51.2 | 67.3 | 60.6 | 70.2 |
| C1 | Ours (Raw-Text) | 46.8 | 64.0 | 42.7 | 61.1 | 37.1 | 57.4 | 47.2 | 56.0 |
| C2 | Ours (w/o SSRE) | 58.0 | 72.4 | 55.1 | 70.2 | 46.3 | 64.1 | 53.5 | 65.4 |
| C3 | Ours (w/o ICA) | 56.2 | 71.5 | 54.0 | 69.3 | 48.0 | 65.0 | 55.2 | 67.2 |
| C4 | Ours | **64.1** | **78.4** | **61.0** | **75.2** | **57.0** | **71.0** | **62.4** | **73.1** |

Table 9 and Table 10 further break down performance by question type. On NuScenes-QA, C4 achieves the best scores on all subtasks, including 67.3% / 80.6% (EM/F1) on Relation QA, 64.2% / 78.1% on Dynamic/Status QA, 60.5% / 74.0% on Risk Prediction QA, and 76.2% on Decision QA. On Waymo-QA, C4 consistently leads with 64.1% / 78.4% on Relation QA, 61.0% / 75.2% on Dynamic/Status QA, 57.0% / 71.0% on Risk Prediction QA, and 73.1% Decision Accuracy. Furthermore, the raw-text baseline (C1) underperforms across all subtasks, highlighting the necessity of structured, coordinated inputs for open-world reasoning.

#### 5.5.3 Ablation study: ICA, SSRE, and IVL

As shown in Table 11, we conduct an ablation study by averaging results over NuScenes-QA and Waymo-QA to isolate the contribution of each component. The full model (C4) yields the best average performance (62.8% QA EM, 76.4% Answer F1, 74.7% DA, 64.4% RI F1) and markedly improves reliability (92.3% FCPR with only 8.0% HER). Introducing ICA alone (C2 vs. C1) substantially improves both correctness and reliability, indicating that converting noisy perception outputs into a clean SceneSummary is a prerequisite for stable reasoning. In contrast, removing IVL (C5) causes a clear drop in reliability (FCPR 87.3% and HER 10.7%), while task correctness changes marginally, suggesting that verifier-style self-checking primarily serves as the final safeguard against hallucination-driven justifications.

Table 11. Ablation results of the ICA and SSRE workflow averaged over the NuScenes-QA and Waymo-QA datasets (%).

| ID | Method | QA EM | Answer F1 | DA | RI F1 | FCPR | HER |
|---|---|---|---|---|---|---|---|
| C1 | Ours (Raw-Text) | 47.6 | 64.7 | 58.2 | 48.9 | 53.8 | 42.8 |
| C2 | Ours (w/o SSRE) | 55.8 | 70.9 | 66.8 | 55.5 | 77.7 | 18.6 |
| C3 | Ours (w/o ICA) | 56.6 | 71.6 | 68.8 | 57.2 | 79.8 | 16.9 |
| C5 | Ours (w/o IVL) | 62.1 | 75.7 | 74.0 | 62.9 | 87.3 | 10.7 |
| C4 | Ours | **62.8** | **76.4** | **74.7** | **64.4** | **92.3** | **8.0** |

#### 5.5.4 Multidimensional capability visualization and generalization gap

To provide an intuitive, holistic view of how different paradigms trade off task correctness, safety, reliability, and explanation quality, we adopt radar charts over six normalized dimensions: QA EM, Decision Accuracy (DA), Risk Identification F1 (RI F1), Fact Consistency Pass Rate (FCPR), 1−HER, and ROUGE-L. Following our experimental protocol, each radar chart overlays five representative curves: VLM Avg (A1–A3), Talk2BEV-style, DriveGPT4-style, PlanAgent-style, and Ours (C4). All metrics are normalized to [0,1] when visualized as absolute scores to ensure comparability across axes.

A direct side-by-side comparison of absolute scores on NuScenes-QA vs. Waymo-QA can become visually insensitive when the two benchmarks are intentionally aligned in task definition and question templates (as is the case for our Waymo-QA construction). To make cross-dataset robustness explicit and directly answer the protocol's generalization question, we adopt a two-panel design:

Left panel ("NuScenes-QA (Absolute, normalized to [0,1])") visualizes the absolute capability profile on NuScenes-QA, capturing the multi-axis performance footprint and paradigm-level trade-offs.

Right panel ("Waymo-QA (Generalization Gap, pp; Outer=Better)") visualizes the generalization gap from NuScenes-QA to Waymo-QA on a per-metric basis. For each metric $m$, we define the gap in percentage points (pp) as:

$$\Delta_m(pp) = Score_m^{NuScenes} - Score_m^{Waymo}.$$

The radar charts yield three consistent observations. First, the full system forms the broadest capability envelope on NuScenes-QA, driven by simultaneous gains on safety and reliability axes rather than sacrificing correctness. Second, explanation similarity and factual reliability are separable: a higher ROUGE-L (e.g., DriveGPT4-style) does not necessarily imply stronger factual consistency or lower hallucination, reinforcing the need to jointly report reliability metrics in safety-critical QA. Third, the generalization-gap view shows that the full system degrades least on reliability axes (e.g., only 1.8 pp drop on FCPR and 1.2 pp drop on 1−HER), while some capability dimensions (e.g., risk) remain more sensitive to domain shift, providing a nuanced picture of robustness beyond aggregate scores.

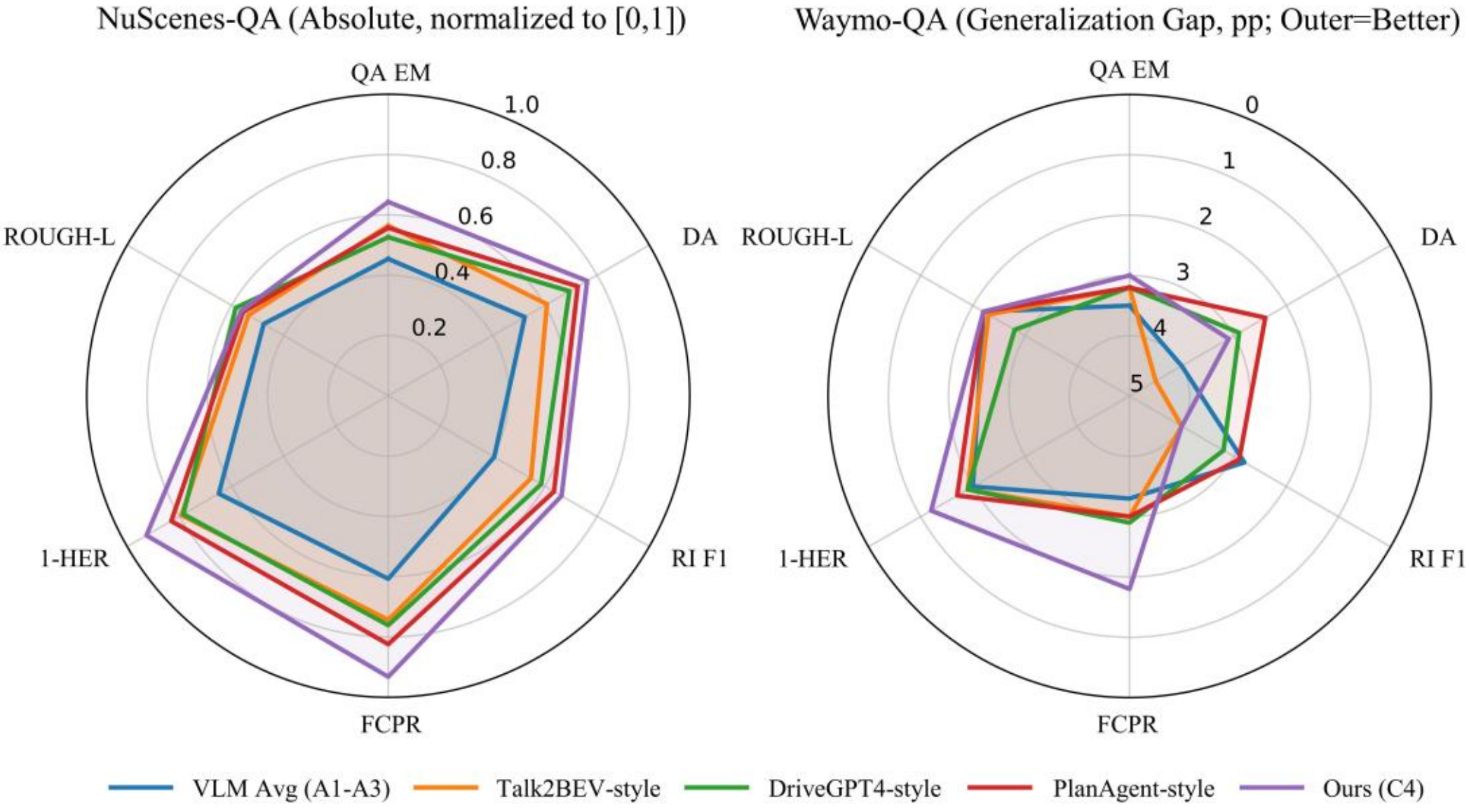

Figure 5. Multidimensional capability radar: absolute performance vs cross-dataset generalization.

### 5.5.5 Summary

Across NuScenes-QA and Waymo-QA, the full ICA+SSRE system (C4) achieves the strongest overall trade-off between correctness, safety, and reliability. It delivers the best QA/decision scores while substantially improving factual grounding (FCPR=93.2% / 91.4%) and keeping hallucinations low (HER=7.4% / 8.6%). Task-type results show consistent gains on relation, dynamic, risk, and decision queries, and the ablation study attributes the improvements to complementary roles: ICA primarily stabilizes grounding by constructing a clean SceneSummary, whereas SSRE—especially IVL—further reduces hallucination-driven inconsistencies with limited impact on raw correctness. Overall, these results support the core claim that a coordinated single-source world model combined with verifier-aware reasoning is critical for dependable open-world driving QA and high-level decision making under domain shift.

### 5.6 Qualitative case studies

This section provides qualitative case studies to complement the quantitative evaluation, illustrating how the proposed neuro-symbolic architecture improves reliability in open-domain driving scene understanding. Using three representative scenes from nuScenes v1.0-mini, we visualize (i) cross-modal conflict resolution and completeness in the intermediate world model (SceneSummary), (ii) disambiguation benefits of a unified structured interface for downstream reasoning, and (iii) verifiable risk-aware decision making under hallucination-inducing prompts.

Figures 6-8 summarize each case with a front-camera view and a LiDAR BEV visualization, followed by two natural-language queries (Q1–Q2) and the corresponding answers from representative baselines and our method. To improve readability, blue highlights mark the key claims that the reader should attend to in our outputs (typically grounded entities/relations/actions that are directly supported by the Evidence box), while red highlights mark salient failure cues in baseline outputs (e.g., unresolved conflicts, explicit uncertainty, or unsupported/incorrect premises). In addition, each case includes a compact "Evidence" box: Cases A–B primarily cite ICA-cleaned SceneSummary records, whereas Case C primarily cites SSRE trace snippets (SRA/VRCG/IVL) that are explicitly grounded on the ICA SceneSummary.

#### 5.6.1 Case study A: cross-modal conflict resolution for safety-critical attributes

Figure 6 shows scene-0061 (sample #10) in nuScenes v1.0-mini, where the ego vehicle approaches an intersection influenced by roadside construction. The scene contains (i) a closest leading vehicle whose semantic type is reported inconsistently across modalities, and (ii) pedestrians near the crosswalk/conflict zone that require conservative driving behavior. We ask: (Q1) the type of the closest leading vehicle ahead (e.g., sedan/SUV/van/MPV), and (Q2) whether pedestrians exist near the conflict zone and what action is recommended.

The Naive Fusion + LLM baseline exposes the core difficulty: when the prompt contains contradictory type reports for the same spatial location, the model tends to either remain explicitly uncertain or produce hedged answers, which weakens its usefulness for downstream decisions. End-to-end V-L agents (DriveGPT4-style) and BEV-based VLMs (Talk2BEV-style) may commit to a single type, but their commitment can differ across methods, reflecting the lack of an explicit, shared conflict-resolution mechanism in the interface used for reasoning in this case.

In contrast, our ICA module resolves the cross-modal type conflict into a single fused entity record and retains a single-sensor pedestrian detection for completeness, producing a clean SceneSummary as the "single source of truth." The answer of ICA+SSRE therefore becomes consistent across Q1–Q2: the leading vehicle type is reported as van/MPV (with the prior cross-modal disagreement auditable), and the presence of pedestrians near the crosswalk triggers a conservative action ("slow down and prepare to yield/stop before entering the conflict zone"). The Evidence box in Figure 6 is intentionally minimal and primarily sourced from ICA output: (i) a fused record with an explicit conflict_resolved field for the lead vehicle, and (ii) a pedestrian record retained despite a radar miss, supporting both correctness and safety-critical completeness.

**Case A: Cross-Modal Conflict Resolution**

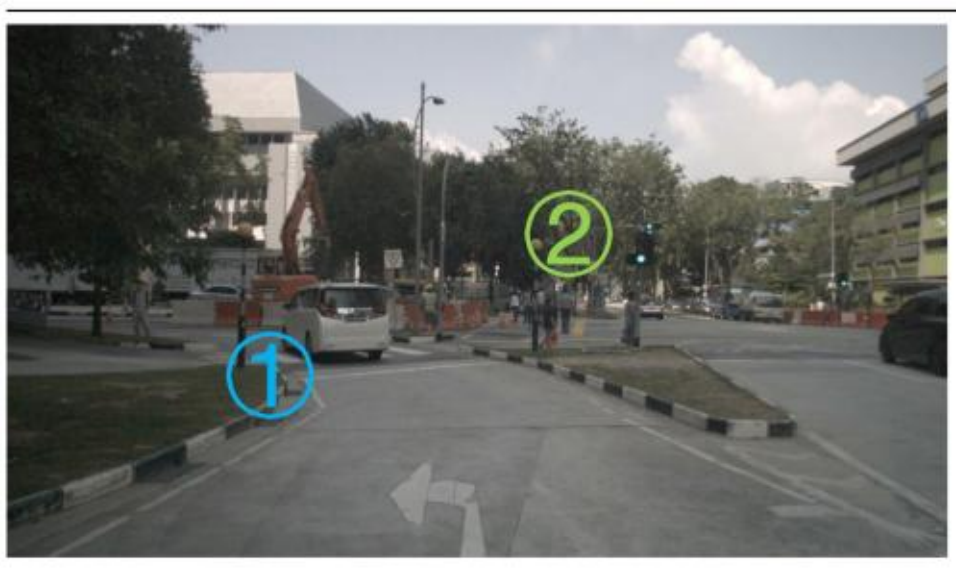

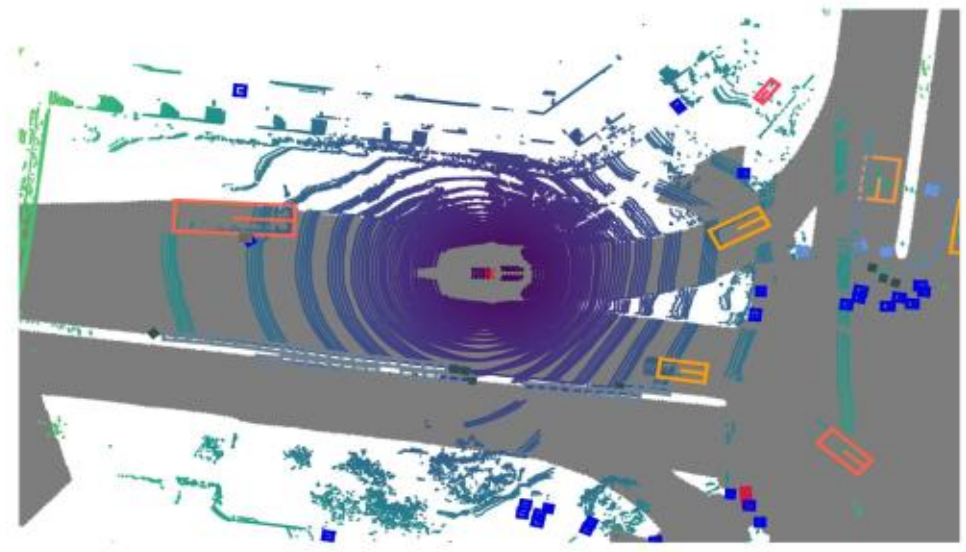

**Q1** (*Conflict-sensitive object attribute*):
What is the type of the closest leading vehicle ahead of the ego vehicle (e.g., sedan/SUV/van/MPV)?
**Q2** (*Missed-detection & safety-critical check*):
Are there pedestrians near the conflict zone (crosswalk / intersection entry) that the ego vehicle must yield to? What is the recommended action?

| | |
|---|---|
| **Naive Fusion** (LLaMA-based) | **A(Q1):** The closest vehicle ahead is described inconsistently across sensors: the camera suggests a van/MPV, while LiDAR reports an SUV at a similar location. These may refer to the same object, so the vehicle type cannot be determined with high confidence.<br>**A(Q2):** Pedestrian detections are not consistent across sensors. There may be pedestrians near the conflict zone, but the evidence is incomplete. **Recommended action:** slow down and be prepared to stop before entering the conflict zone. |
| **DriveGPT4-style** (E2E V-L Agent) | **A(Q1):** A white SUV is directly ahead, moving slowly near the construction zone.<br>**A(Q2):** Pedestrians are present near the crosswalk. **Recommended action:** brake early, maintain a safe gap, and be ready to yield/stop. |
| **Talk2BEV-style** (BEV-based VLM) | **A(Q1):** The closest leading vehicle ahead is likely a white van/MPV in the ego direction of travel.<br>**A(Q2):** Pedestrians are near the conflict zone. **Recommended action:** slow down and be ready to yield before the crosswalk. |
| **Ours** (ICA+SSRE) | **A(Q1):** The closest leading vehicle ahead is a white van/MPV, moving slowly in front of the ego vehicle.<br>**A(Q2):** Yes—pedestrians are present near the conflict zone (crosswalk vicinity). **Recommended action:** slow down and prepare to yield/stop before entering the conflict zone; proceed only after the crossing area is clear. |

**Evidence (Primary: ICA SceneSummary, after conflict resolution):**

**1. Lead vehicle (conflict resolved → single truth)**

{"entity_id":"Veh_12","entity_type":"van/mpv","source_modalities":["camera","lidar"],"conflict_resolved":{"camera":"van/mpv","lidar":"SUV"}}

**2. Pedestrian near crosswalk (single-source retained for completeness)**

{"entity_id":"Ped_03","entity_type":"pedestrian","zone":"crosswalk_vicinity","source_modalities":["camera"],"note":"single-source (radar missed)"}

Figure 6. Case A: Cross-modal conflict resolution for safety-critical attributes.

### 5.6.2 Case study B: Motion–intent disambiguation with a unified SceneSummary interface

Figure 7 presents scene-0103 (sample #4), a typical urban intersection where fine-grained motion status and turning intent must be inferred to reason about conflict-zone occupancy. We query: (Q1) whether the closest ego-lane vehicle ahead is moving or stationary, and (Q2) what the front-left black car is doing (turning/going straight/parked) and whether it will enter the ego vehicle's conflict zone.

This case highlights an interface-driven ambiguity: single-frame image evidence (LLaVA) often provides limited observability for motion, while text-only reasoning on uncoordinated multi-sensor logs (Naive Fusion + LLM) may be dominated by duplicated or inconsistent entries, resulting in low-confidence answers about "moving vs. stopped" and "will enter vs. will not enter." BEV-based VLMs (Talk2BEV-style) typically offer clearer spatial judgments, yet the absence of an explicit, typed world state can still leave intent and conflict-zone relations under-specified.

Our method and evidence. With ICA, the SceneSummary explicitly encodes lane_relation, motion_state, intent, and zone for the relevant entities, turning the ambiguous “interpretation problem” into a direct “query over a structured world model.” Consequently, ICA+SSRE answers Q1–Q2 with consistent, decision-relevant statements: the closest ego-lane vehicle is moving, and the front-left black car is turning left and will temporarily enter the conflict zone; the recommended action is to slow down and be prepared to yield/stop before the crosswalk until the conflict clears. Here, the Evidence box is again primarily sourced from ICA output (SceneSummary fields), because the key contribution in this case is disambiguation via a unified interface rather than a multi-step verifier.

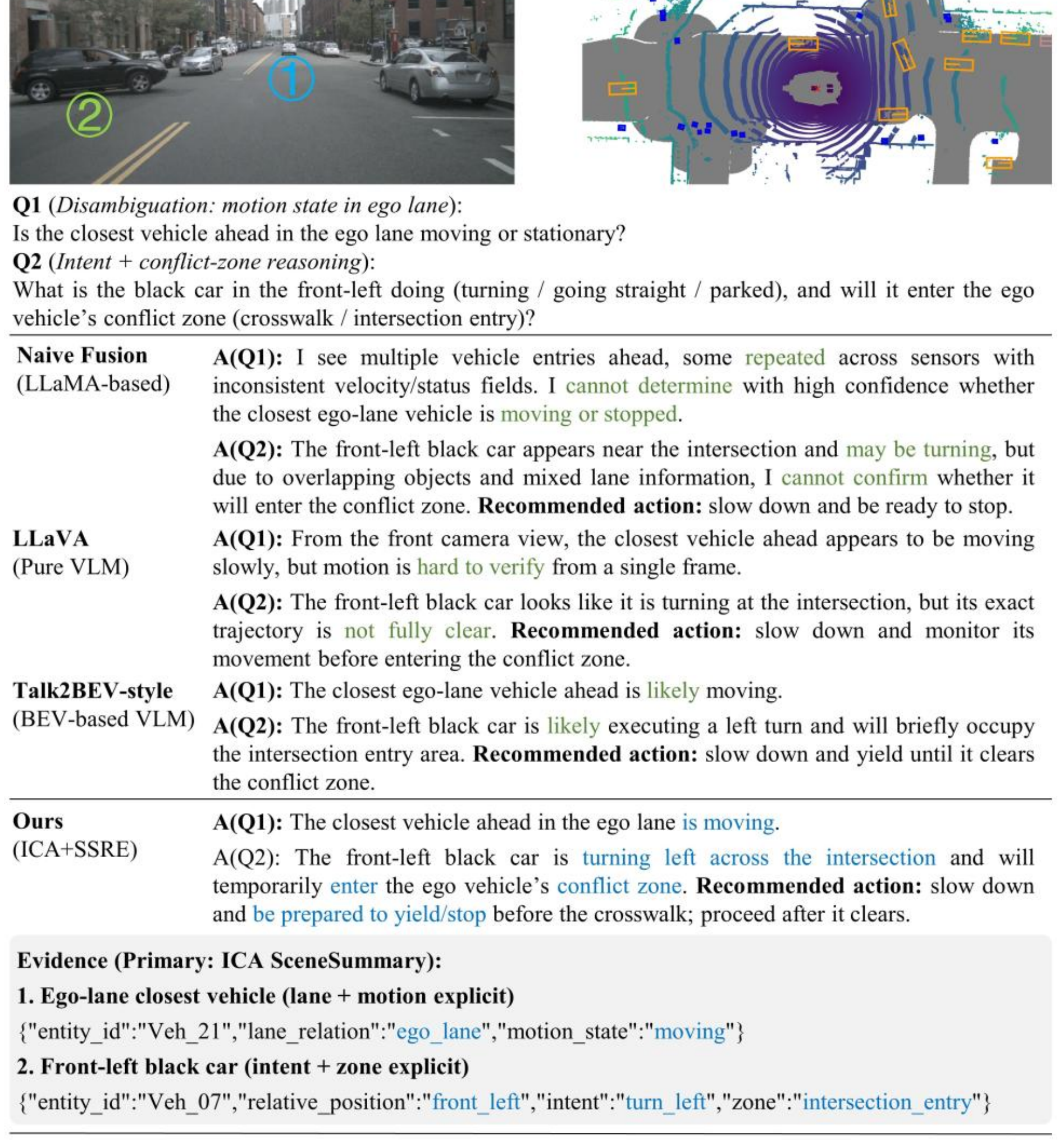

Figure 7. Case B: Motion–intent disambiguation with a unified SceneSummary interface.

### 5.6.3 Case study C: Verifiable risk reasoning under false-premise prompts

Figure 8 shows scene-1094 (sample #27), a nighttime driving scenario with reduced visibility and vulnerable road users near the roadside. Beyond normal risk-aware decision making (Q1), we include a “false-premise” query (Q2) that explicitly asks about a bicycle overtaking on the right, which is a common trigger for hallucinated entities in generative models.

**Case C: Verifiable Risk Reasoning under False-Premise Attacks**

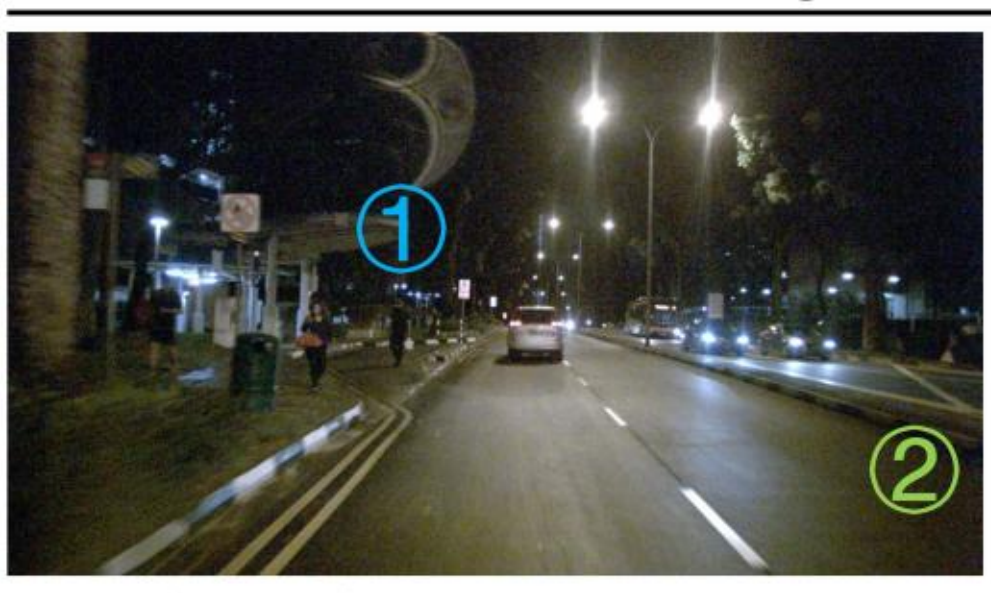

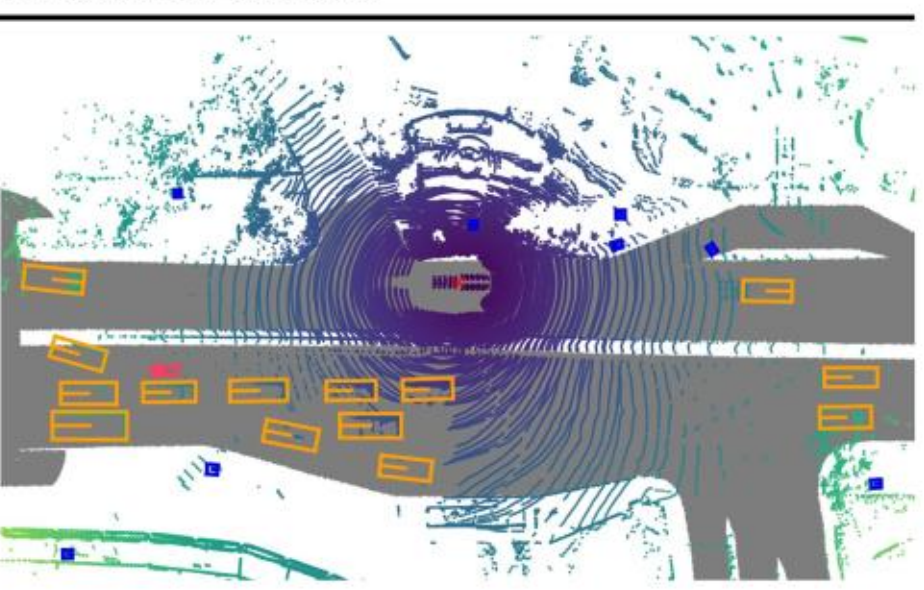

**Q1** (*Risk-aware decision*):
What are the primary risks in this nighttime scene, and what is the recommended driving action (keep lane / slow down / stop / change lane)?
**Q2** (*Hallucination stress test: false premise* ):
Is there a bicycle overtaking on the right? If yes, what should the ego vehicle do?

| | |
|---|---|
| **LLaVA** (Pure VLM) | **A(Q1):** It is nighttime with limited visibility. I see a lead vehicle ahead and possible pedestrian activity near the roadside. **Recommended action:** slow down and keep lane, staying ready to brake.<br>**A(Q2):** I cannot clearly confirm a bicycle from this view; there might be a two-wheeled object on the right. **Recommended action:** keep lane and slow down slightly, leaving extra margin while observing. |
| **PlanAgent-style** (structured planning agent) | **A(Q1):** Plan: (1) reduce speed due to low visibility; (2) increase following distance to the lead vehicle; (3) monitor pedestrians near the roadside; (4) prepare for emergency braking. **Recommended action:** slow down and keep lane.<br>**A(Q2):** The observation does not reliably confirm a bicycle. If a bicycle is overtaking on the right, **recommended action:** maintain lane discipline, reduce speed, and allow it to pass safely (avoid sudden steering). |
| **DriveGPT4-style** (E2E V-L Agent) | **A(Q1):** Nighttime driving requires extra caution. I see a lead vehicle and roadside pedestrians. **Recommended action:** slow down, keep lane, and prepare to stop.<br>**A(Q2):** Yes, a bicycle is overtaking on the right. **Recommended action:** hold your lane, slow down, and give it room to pass safely. |
| **Ours** (ICA+SSRE) | **A(Q1):** Primary risks include reduced visibility and vulnerable road users near the roadside, plus maintaining a safe gap to the lead vehicle. **Recommended action:** slow down and keep lane, increase following distance, and stay ready to brake.<br>**A(Q2):** Premise not supported by the consistent world model: no bicycle entity is present in the SceneSummary for this frame. **Recommended action:** do not take avoidance actions for a non-existent target—keep lane and continue observing. |

**Evidence (Primary: SSRE trace; grounded on ICA SceneSummary):**
**1. SRA top risks (risk → entity anchoring)**
{"sra_top_risks":[{"risk":"VRU near roadside in low light","entity_ids":["Ped_02"],"severity":"High"}]}
**2. VRCG final decision JSON (action → supporting entities)**
{"recommended_action":"slow_down_keep_lane","supporting_entity_ids":["Veh_10","Ped_02"],"confidence":0.70}
**3. ICA SceneSummary check (negative evidence)**
{"entity_types_present":["vehicle","pedestrian"],"entity_types_absent":["bicycle"]}
**4. IVL verification (false premise rejected)**
{"ivl_check":{"premise":"bicycle overtaking on right","supported":false,"resolution":"reject_premise_keep_lane_observe"}}

Figure 8. Case C: Verifiable risk reasoning under false-premise prompts.

Pure VLM reasoning (LLaVA) and structured planning-style agents (PlanAgent-style) often respond cautiously to Q2 (e.g., “cannot clearly confirm”), but may still condition actions on an unverified premise (“if a bicycle is overtaking”), which can be acceptable for conservative driving yet remains evidentially weak. In contrast, the DriveGPT4-style agent in this case explicitly accepts the false premise and asserts the existence of a bicycle, exemplifying an unsafe failure mode: recommending avoidance behavior for an entity that is not supported by the perceived world state.

Our method and evidence. ICA+SSRE handles Q2 by explicitly separating "question premise verification" from "action generation." The SSRE trace first ranks risks via SRA using grounded entity references, then produces a VRCG decision with supporting_entity_ids, and finally invokes IVL to verify whether the bicycle premise is supported by the SceneSummary. Since the consistent world model contains no bicycle entity type, IVL rejects the premise and the final recommendation avoids unnecessary avoidance maneuvers ("do not take avoidance actions for a non-existent target; keep lane and continue observing"). Accordingly, the Evidence box in Figure 8 is primarily sourced from SSRE traces grounded on the ICA SceneSummary: an SRA risk item linked to pedestrian entities, a VRCG decision JSON with explicit supporting_entity_ids, a negative type-presence check from the SceneSummary, and an IVL verdict that marks the premise as unsupported.

### 5.6.4 General discussion for case study

Across the three cases, a consistent pattern emerges: reliability gains come primarily from controlling the information interface and verifiability constraints, rather than relying on the generative model to reconcile noisy evidence on its own. Cases A–B show that ICA's deterministic coordination turns multi-sensor disagreements and omissions into a clean, typed SceneSummary with auditable provenance, enabling downstream models to answer attribute, motion, and intent questions without being dominated by cross-modal redundancy or contradictions.

Case C further indicates that a clean world model alone is insufficient under adversarial or misleading queries; a verifier-aware reasoning workflow is required to prevent premise-driven hallucinations from propagating into actions. By grounding each decision in supporting entity IDs and enforcing an explicit verification step (IVL), SSRE makes it possible to (i) reject unsupported premises, (ii) keep risk assessment aligned with actual entities, and (iii) provide concise, audit-friendly evidence traces.

These qualitative findings align with the quantitative trends reported earlier (higher consistency and lower hallucination rates), reinforcing the central thesis of this work: a structured single-source world representation (SceneSummary) plus verifiable reasoning constraints (SSRE) is a practical design for reliable autonomous driving scene understanding.

## 5.7 Overall discussion

Across Sections 5.2–5.6, the results support the main claim of this work: reliable open-domain driving cognition is achieved by (i) coordinating heterogeneous perception into a single auditable world state and (ii) constraining reasoning with verifiability, rather than expecting a generative model to reconcile noisy, conflicting evidence implicitly. In our pipeline, ICA constructs a globally consistent SceneSummary, and SSRE (with IVL) enforces risk-aware and self-verifying reasoning on top of that representation.

At the perception stage, ICA maintains competitive detection accuracy while substantially improving fusion usability. On nuScenes/Waymo, Ours achieves 70.9% / 73.2% mAP/NDS and 69.8% mAP yet produces a far cleaner fused output (ERR=0.7% / 1.0%, ACR=98.0% / 95.0%). Without coordination, naïve aggregation (Ours w/o ICA) yields severe redundancy and inconsistency (ERR=25–30%, ACR=70–75%). This pattern is reinforced by the scene-consistency benchmark: ICA-based fusion reaches CRR=91.7% and MDCR=74.6% with only HE=0.1 hallucinated entities per scene, whereas representative perception-to-LLM pipelines remain at CRR < 5% and introduce 0.9–1.6 hallucinations per scene.

Section 5.4 shows that the unified SceneSummary interface benefits LLM reasoning in both effectiveness and efficiency. Replacing raw multi-agent detection logs (~1000 tokens) with a compact SceneSummary-plus-synopsis (~260 tokens) improves Accuracy by 20–25 points and reduces HR to 5–10% across backbones, while cutting APL by ~74% (~3.8×) and reducing AIT by ~36–42%. Notably, these results indicate that input structure/quality can partially offset model scale, which supports using a typed interface rather than relying on larger models or ad-hoc prompting alone.

Section 5.5 provides end-to-end evidence on open-world QA and decision reasoning under domain shift. The full ICA+SSRE system performs best on both NuScenes-QA and Waymo-QA (e.g., 64.3% / 77.5% QA EM/Answer F1 and 76.2% DA on NuScenes-QA; 61.3% / 75.2% and 73.1% on Waymo-QA) and achieves strong grounding (FCPR=93.2% / 91.4%) with low hallucination (HER=7.4%/8.6%). The ablation study clarifies complementary roles: ICA primarily stabilizes the factual basis (C2 vs C1: FCPR 53.8% to 77.7%, HER 42.8% to 18.6%), while IVL mainly strengthens reliability (C5 vs C4: FCPR 87.3% vs 92.3%, HER 10.7% vs 8.0%) with marginal changes to raw correctness. Moreover, the cross-dataset gaps on reliability remain small (ΔFCPR=1.8 pp; Δ(1−HER)=1.2 pp), and the radar analysis emphasizes that BLEU/ROUGE-style overlap does not substitute for factual faithfulness—motivating explicit reporting of FCPR/HER in safety-critical evaluation.

The qualitative case studies in Section 5.6 corroborate these quantitative trends with mechanism-level evidence. Cases A–B show that ICA converts cross-modal disagreements and omissions into a typed, auditable SceneSummary that supports consistent answers for safety-critical attributes and motion/intent queries. Case C further highlights robustness under misleading prompts: by separating premise verification from action generation, SSRE+IVL rejects unsupported entities (e.g., the queried bicycle) and prevents premise-driven hallucinations from propagating into high-level decisions.

## 6. Conclusion

This paper presented InfoCoordiBridge, a BEV-centric neuro-symbolic architecture for reliable open-world scene understanding in autonomous driving. The proposed framework bridges multimodal perception and large language model (LLM) reasoning through an explicit coordination layer. Concretely, we introduced (i) a unified multi-agent perception interface that produces well-typed structured facts and modality-focused synopses, (ii) an Information Coordination and Abstraction (ICA) module that deterministically aligns entities, resolves cross-modal conflicts, and constructs an auditable SceneSummary as a single source of truth, and (iii) a SceneSummary-Driven Reasoning Engine (SSRE) that performs verifiable chain-of-prompt reasoning with explicit entity grounding and an introspective verification loop.

Extensive experiments on nuScenes and Waymo demonstrate that the proposed approach achieves competitive 3D detection performance while substantially improving fusion consistency and reliability. In particular, our method reaches 70.9% mAP / 73.2% NDS on nuScenes and 69.8% mAP on Waymo, and the ICA module yields low inconsistency error rates (e.g., ERR=0.7% / 1.0% and ACR=98.0% / 95.0% on nuScenes/Waymo) while maintaining strong detection accuracy. Moreover, on open-world QA and decision reasoning, the full system achieves the best overall correctness and reliability, reaching QA EM=64.3% / 61.3%, Answer F1=77.5 / 75.2, and DA=76.2 / 73.1 on NuScenes-QA and Waymo-QA, respectively, while attaining high fact-consistency (FCPR=93.2% / 91.4%) and low hallucination rates (HER=7.4% / 8.6%). These results support the central thesis that coordinating perception into a clean symbolic world state before invoking LLM reasoning is an effective way to mitigate hallucinations and improve robustness by construction.

Future work will focus on extending the framework to longer temporal horizons and more interactive multi-agent traffic settings, improving efficiency for real-time deployment, and strengthening formal safety guarantees (e.g., tighter verification objectives and richer consistency constraints). Another promising direction is to reduce reliance on prompt engineering by learning compact, task-aware verifiers and by integrating the SceneSummary representation more tightly with downstream planning and control.

**CRediT authorship contribution statement**

**Shuo Liu:** Writing – original draft, Methodology. **Lei Shi:** Conceptualization. **Haowen Liu:** Resources, Data curation. **Jing Xu:** Formal analysis, Supervision. **Yufei Gao:** Investigation, Visualization. **Yucheng Shi:** Writing – review and editing, Conceptualization.

**Declaration of competing interest**

The authors declare that they have no known competing financial interests or personal relationships that could have appeared to influence the work reported in this paper.

**Data availability**

Data will be made available on request.

**Acknowledgements**

This work was supported in part by the National Natural Science Foundation of China (62406293), the Key research and development program of Henan Province (251111212100) and the Postdoctoral Innovative Talent Support Program (GZC20251644).

**Appendix A. Pseudocode of the Information Coordination and Abstraction (ICA) module**

The ICA module deterministically aligns multi-agent perception outputs into a unified, conflict-resolved world state called SceneSummary. In our setting, LiDAR/BEVFusion/Radar provide geometric estimates in the ego-centric BEV frame, whereas the camera agent mainly provides fine-grained semantic attributes (e.g., color and intent cues) from 2D observations. Importantly, the association between camera observations and BEV entities is performed via projection-based matching and semantic compatibility, without assuming that the dataset provides per-pixel depth.

Algorithm A.1 ICA pipeline (compact pseudocode)

**Inputs:**

- D^B: BEVFusion detections (3D boxes / states in ego-BEV)
- D^L: LiDAR detections (3D boxes / states in ego-BEV)
- D^R: Radar detections (2D/3D states in ego-BEV, depending on sensor)
- D^C: Camera detections (2D boxes + semantics in image; optional camera-only 3D if available)

- Calib: camera intrinsics/extrinsics for projection
- Θ : gating thresholds, matching thresholds, modality priors and reliabilities

**Output:**

- S: SceneSummary (a single, conflict-resolved structured world state)

**Procedure:**

1) Normalize and wrap agent outputs

For each modality s ∈ {B, L, R}:

- Convert detections to a common ego-centric BEV state form:

z_i^s = (μ_i^s, Σ_i^s, class_i^s, attrs_i^s, score_i^s)

For camera modality C:

- Keep 2D boxes b_k^C and semantic attributes attrs_k^C (e.g., color, intent cues).
- Do NOT require depth; geometry is not updated from camera-only 2D observations.

2) Stage-1 association: LiDAR ↔ BEVFusion (build geometric seeds)

- Build candidate pairs via gating in BEV (e.g., distance / Mahalanobis).
- Compute a composite matching cost that prioritizes geometry consistency.
- Solve one-to-one assignment (Hungarian) to obtain matches M_{L,B}.
- Initialize entity seeds E from matched pairs; keep unmatched detections as single-source entities.

3) Stage-2 association: Radar → Seeds (motion refinement)

- Build candidate pairs between radar detections and seeds via gating and velocity consistency.
- Solve assignment and attach radar observations to the corresponding entities.

4) Stage-3 association: Camera semantics → Entities (projection-based attachment)

For each camera view:

- Project each entity's current 3D box/state to the image plane using Calib, obtaining a predicted 2D region \hat(Bai, et al.)(e).
- For each camera detection b_k^C, compute IoU(\hat(Bai, et al.)(e), b_k^C) and check class compatibility.
- Solve assignment between entities and camera detections using an IoU-dominant cost.
- For each matched pair (e, k):

attach attrs_k^C to entity e as semantic evidence (no geometry overwrite).

5) Conflict checking and attribute fusion (per entity)

For each entity e:

- Collect all available observations {z^s} from its attached modalities.
- For continuous attributes (position, velocity, size, yaw):

fuse using modality-aware information weighting;

if sources are potentially dependent, apply covariance intersection (CI).

- For discrete attributes (class, color, intent tags):

fuse via weighted voting with modality reliability priors.

- Produce ambiguity flags when evidence remains inconsistent.

6) Structured summary generation

- Serialize the fused entities into SceneSummary S with:

id, type/class, fused state, confidence, source lineage, ambiguity flags, and optional scene-level notes.

Return S.

## Appendix B. Prompt templates for the chain-of-prompt reasoning workflow

This section lists the prompts used in SSRE (Section 4.3) to support structured parsing, systematic risk assessment, decision reasoning, introspective verification, and revision. All prompts treat the *SceneSummary* produced by ICA as the single source of truth for the current reasoning instance, and they prohibit introducing new entities that do not appear in the input.

### B.1. Prompt template for structured scene parsing ($T_parse$)

T_parse | Structured Scene Parsing

**Role:** You are a scene-graph parser for autonomous driving.

**Input:**

- SceneSummary S (JSON), containing a list of entities with IDs and state estimates.

**Output:**

- M (a structured intermediate representation) in the exact schema below.

**Rules:**

1) Do not invent entities, attributes, or relations.

2) You may compute ONLY deterministic derived quantities from S (e.g., speed magnitude, pairwise distance, relative direction, and "approaching" based on position/velocity). Do not infer intentions unless S explicitly contains an intent tag.

3) Use entity references strictly in the form ID_<number> (e.g., ID_17).

4) If a field is missing in S, output "unknown" for that field.

**Output schema** (YAML-like, keep the keys and order):

ENTITIES:

- id: ID_17
  type: <from S>
  position_bev_m: [x, y]
  velocity_bev_mps: [vx, vy]
  size_m: [l, w, h]
  heading_rad: <value or unknown>
  confidence: <value or unknown>
  sources: <list>
  semantic_attributes: <list or empty>
  flags: <list or empty>

RELATIONS (optional; only if computable from S):

- subject: ID_17
  relation: <ahead_of | behind_of | left_of | right_of | near | far | approaching | moving_away>
  object: ID_35
  evidence: <brief numeric evidence based on S, e.g., distance=8.2m>

Now parse S into M.

## B.2. Prompt template for systematic risk assessment (*T_risk*)

---

T_risk | Systematic Risk Assessment

---

**Role:** You are a driving safety risk assessment agent.

**Input:**

- M (structured scene representation)

**Output:**

- L_risk: a prioritized JSON array of risk items, sorted by (severity, urgency) descending.

**Rules:**

1) Base the assessment ONLY on M. Do not assume unobserved traffic rules, signals, or occlusions unless stated.

2) Each item must reference involved entities using ID_<number>.

3) Use severity and urgency on a 1–5 scale (5 = highest).

4) If there is insufficient evidence, output a risk with low confidence rather than fabricating details.

**Output JSON schema**:

```
[
  {
    "risk_type": "<collision | cut_in | rear_end | pedestrian_crossing | occlusion | rule_violation | unknown>",
    "involved_entity_ids": ["ID_17", "ID_35"],
    "severity": 1-5,
    "urgency": 1-5,
    "confidence": 0.0-1.0,
    "evidence": "A short, verifiable statement that cites fields/relations from M."
  },
  ...
]
```

Now generate L_risk.

---

## B.3. Prompt template for decision reasoning (*T_reason*)

---

T_reason | Decision Reasoning with Grounded Justification

---

**Role:** You are an autonomous driving decision agent that must remain strictly grounded.

**Inputs:**

- User query Q (natural language)
- M (structured scene representation)
- L_risk (prioritized risk list)
- Optional auxiliary descriptions A (free-text; may be noisy)

**Outputs:**

1) D_draft (JSON decision)

2) N_draft (natural-language justification)

**Rules:**

1) Treat M as the primary source of truth. A can be used only to provide linguistic phrasing and MUST NOT introduce new entities or contradict M.

2) Every factual claim in N_draft must cite at least one entity ID in parentheses, e.g., "(ID_17)".

---

3) If evidence is insufficient, choose a conservative action and lower the confidence.

4) Do not output hidden reasoning traces; provide a concise, verifiable justification.

**Decision JSON schema**:

{
  "recommended_action": "<keep_lane | slow_down | stop | yield | change_lane_left | change_lane_right | follow | unknown>",
  "confidence": 0.0-1.0,
  "target_entity_ids": ["ID_17", "..."],
  "risk_summary": "One sentence summarizing the main risks considered.",
  "constraints": ["optional safety constraints or checks, if any"]
}

Now answer Q using (M, L_risk, A) and output D_draft first, then N_draft.

### B.4. Prompt template for introspective verification ( *T_verify* )

T_verify | Introspective Verification

**Role:** You are a verifier. Your task is to check whether the draft decision and justification contradict the SceneSummary S.

**Inputs:**

- SceneSummary S (JSON)
- D_draft (decision JSON)
- N_draft (justification text)

**Output:**

- V (a JSON object)

**Verification checklist:**

1) No new entity IDs appear in D_draft or N_draft that are absent in S.

2) Any stated entity type/state (position/velocity/size) must not contradict S.

3) If a claim cannot be verified from S, mark it as unsupported.

4) Focus on factual grounding, not writing style.

**Output JSON schema:**

{
  "verdict": "PASS" | "FAIL",
  "unsupported_or_conflicting_claims": [
    {"claim": "...", "reason": "..."}
  ],
  "comment": "One short comment for revision."
}

Now verify and output V.

### B.5. Prompt template for revision ( *T _revision* )

T_revision | Revision under Verifier Feedback

**Role:** You are a reviser. You must revise (D_draft, N_draft) based on verifier output V.

**Inputs:**

- M, L_risk, Q
- D_draft, N_draft

- V (verifier JSON)

**Outputs:**

- D_final (JSON decision)
- N_final (justification text)

**Rules:**

1) Remove or correct any unsupported/conflicting claims listed in V.
2) Keep entity references consistent (ID_<number>).
3) If conflicts cannot be resolved, choose a more conservative action and reduce confidence.

Now produce D_final and N_final.

## Appendix C. Template design for generating driving QA pairs from the Waymo dataset

To evaluate open-world question answering and decision consistency under domain shift, we convert Waymo scenes into a driving QA format by instantiating rule-based question templates over the scene graph. We provide four representative template families: relation and relative position QA, counting QA, risk prediction QA, and decision-making QA.

We use the ego-centric BEV frame (x-forward, y-left). Relations such as "front/behind/left/right/near" are instantiated by deterministic thresholds on relative coordinates and Euclidean distance. Risk labels can be generated using simple kinematic criteria (e.g., time-to-collision) consistent with the annotations available in Waymo.

### C.1. Template family 1: relation and relative position (Yes/No)

**Question:**

"Is there a <REL> relationship between the <OBJ_1> and the <OBJ_2> in the current scene?"

Examples of <REL>: in front of, behind, to the left of, to the right of, near.

**Answer:**

"Yes" if the instantiated relation holds under the BEV-threshold rules; otherwise "No".

### C.2. Template family 2: counting (integer)

**Question:**

"How many <OBJ_2> are <REL> the <OBJ_1> in the current scene?"

**Answer:**

"<N>", where N is the number of objects satisfying the relation under the BEV-threshold rules.

### C.3. Template family 3: risk prediction (Yes/No + short rationale)

**Question:**

"Is there a potential collision risk between the ego vehicle and the <OBJ_2> in the next <T> seconds?"

**Answer:**

"Yes" if a simple kinematic criterion (e.g., TTC < τ) is triggered; otherwise "No".

(Optionally provide a short rationale: distance and closing speed.)

### C.4. Template family 4: decision-making (multiple-choice)

**Question:**

"Given the current scene, what is the safest immediate action for the ego vehicle?"

**Options:**

A) keep lane and maintain speed

B) slow down

C) stop / yield

D) change lane left

E) change lane right

**Answer:**

Select the option consistent with the generated risk label and scene constraints (e.g., choose B/C when high collision risk is present).